%% file: main.tex
\newcommand{\paperTitle}{Geometric Inconsistency Localization in Multi-View Image Sets}

\documentclass[sigconf, nonacm]{acmart}

\input{preamble.tex}

\begin{document}

\input{header.tex}

\begin{abstract}
\input{abstract.tex}

\end{abstract}

%% Keywords
\keywords{Image Forensics, Geometric Inconsistency Localization, Wide-Baseline Multi-View Image Sets}

%% Maketitle
\maketitle

%%%%%%%%%%%%%%%%%%%%%%%%%%%%%%%%%%%%%%%%%%%%%%%%%%%%%%%%%%%%%%%%%%%%%%%%

\input{body.tex}

%%%%%%%%%%%%%%%%%%%%%%%%%%%%%%%%%%%%%%%%%%%%%%%%%%%%%%%%%%%%%%%%%%%%%%%%

\input{acknowledgments.tex}

%%%%%%%%%%%%%%%%%%%%%%%%%%%%%%%%%%%%%%%%%%%%%%%%%%%%%%%%%%%%%%%%%%%%%%%%

%% References
\bibliographystyle{ACM-Reference-Format}
\bibliography{references}

\end{document}

%% file: preamble.tex
\usepackage{subcaption}
\usepackage{makecell}  % for \makecell -> for line breaks in table cells
\usepackage{multirow}  % for multirow in tables
\usepackage{fp} % Safe substitute for pgf
\usepackage{cleveref}

\newcommand{\titlespace}{\hspace{0.2cm}}

\newcommand{\dottedline}{\addlinespace} %% ACM TAPS safe
\newcommand{\tr}[1]{%
    \FPeval{\result}{round(#1:3)}% 
    \result% 
}

%% file: header.tex
% \setcopyright{acmcopyright} % change as needed
%% Rights management information. This is required.
\copyrightyear{2026}
\acmYear{2026}
% \setcopyright{cc}
\setcctype{by}
\acmConference[DFF '26]{2nd Deepfake Forensics Workshop: Detection, Attribution, Recognition, and Adversarial Challenges in the Era of AI-Generated Media}{November 10--14, 2026}{Rio de Janeiro, Brazil}
\acmBooktitle{2nd Deepfake Forensics Workshop: Detection, Attribution, Recognition, and Adversarial Challenges in the Era of AI-Generated Media (DFF '26), November 10--14, 2026, Rio de Janeiro, Brazil}
\acmDOI{10.1145/3840473.3840513}
\acmISBN{979-8-4007-2927-0/2026/11}

%% Title
\title[\paperTitle]{\paperTitle}

%% Authors
\author{Xander Staelens}
\orcid{0009-0001-4391-4377}
\affiliation{
  \institution{IDLab, Ghent University - imec}
  \city{Ghent}
  \country{Belgium}
}
\email{Xander.Staelens@ugent.be}

\author{Albéric Loos}
\orcid{0009-0002-1140-656X}
\affiliation{
  \institution{IDLab, Ghent University - imec}
  \city{Ghent}
  \country{Belgium}
}
\email{Alberic.Loos@ugent.be}

\author{Bert Ramlot}
\orcid{0009-0006-7787-4218}
\affiliation{
  \institution{IDLab, Ghent University - imec}
  \city{Ghent}
  \country{Belgium}
}
\email{Bert.Ramlot@ugent.be}

\author{Hannes Mareen}
\orcid{0000-0002-0660-3190}
\affiliation{
  \institution{IDLab, Ghent University - imec}
  \city{Ghent}
  \country{Belgium}
}
\email{Hannes.Mareen@ugent.be}

\author{Peter Lambert}
\orcid{0000-0001-5313-4158}
\affiliation{
  \institution{IDLab, Ghent University - imec}
  \city{Ghent}
  \country{Belgium}
}
\email{Peter.Lambert@ugent.be}

\author{Glenn Van Wallendael}
\orcid{0000-0001-9530-3466}
\affiliation{
  \institution{IDLab, Ghent University - imec}
  \city{Ghent}
  \country{Belgium}
}
\email{Glenn.VanWallendael@ugent.be}

%% file: abstract.tex
% \todo{abstract}

Novel view synthesis (NVS) models can produce realistic new views of the same scene from different viewpoints. 
However, these generated views are not always geometrically consistent with one another.  
Multi-view (MV) consistency has shown promise as a tool for evaluating these NVS models.
Its potential for multimedia forensics, however, remains largely unexplored, particularly for localizing geometric inconsistencies across wide-baseline image pairs.
%
% To enable research in this direction, we introduce DeformView, the first MV dataset with pixel-level annotations of geometric inconsistencies in wide-baseline MV image pairs.
To enable research in this direction, we introduce DeformView, a wide-baseline MV dataset with pixel-level annotations of geometric inconsistencies.
Using DeformView, we evaluate state-of-the-art MV consistency-scoring methods and show that approaches developed for NVS evaluation transfer poorly to the forensic task of geometric inconsistency localization.
To address this limitation, we propose DEFECt3R, a lightweight learning-based classifier that uses cross-view feature relationships to localize geometric inconsistencies at the pixel level. 
By learning from explicit supervision, including hard negatives from geometrically consistent yet deformed views, DEFECt3R improves localization performance and substantially reduces false positives compared to existing consistency-scoring methods.
Ablation experiments further show that both feature representations and correspondence quality contribute to localization performance.
%
% Overall, our findings demonstrate that MV geometric consistency is a promising yet underexplored signal for multimedia forensics. 
% % Together, DeformView and DEFECt3R establish the first benchmark and baseline for geometric inconsistency localization in MV image sets.
% %% REPLACED AFTER REVIEW
% Together, DeformView and DEFECt3R establish a benchmark and baseline for geometric inconsistency localization in wide-baseline MV image sets.
%
% FUSED INTO ONE SENTENCE
Overall, our findings demonstrate that MV geometric consistency is a promising yet underexplored signal for multimedia forensics and establish a benchmark and baseline for geometric inconsistency localization in wide-baseline MV image pairs.
Code and dataset are available at \url{https://github.com/IDLabMedia/DeformView-DEFECt3R}

%% file: body.tex
%-----------------------------%
%         INTRODUCTION
%-----------------------------%

\begin{figure}[h]%[tb]
    \centering

    \includegraphics[width=0.95\linewidth]{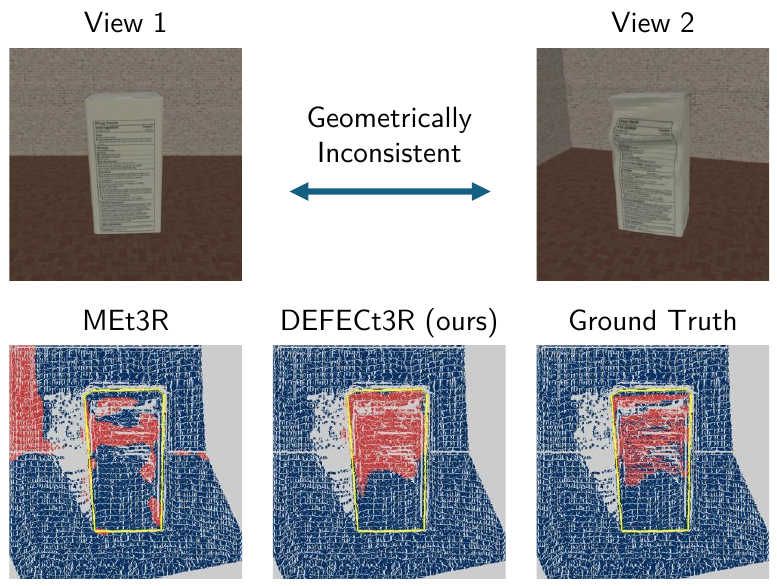} % Drop extension so TAPS can chose PDF for PDF and PNG from HTLM
    
    \caption{
        An image pair, from the proposed DeformView dataset, shows two views of the same object taken from different viewpoints. While both images appear plausible in isolation, they are geometrically inconsistent with one another. 
        The figure shows the responses produced by MEt3R~\cite{MEt3R}, the inconsistency map predicted by our proposed method, DEFECt3R, and the ground-truth annotation.
        Existing geometric consistency methods either produce only a global consistency score or require more views than are typically available in forensic settings. In contrast, DEFECt3R directly localizes geometric inconsistencies, producing pixel-level inconsistency maps that improve explainability while achieving increased detection performance.
    }

    \Description{
        The figure is organized into two rows. The first row shows an image pair of the same object captured from different viewpoints. The second row contains three columns. The first shows a response map produced by MEt3R, where brighter regions indicate higher predicted inconsistency. The second shows the pixel-level inconsistency map predicted by DEFECt3R. The third shows the ground-truth inconsistency mask. The inconsistent region appears as a localized area on the object and is highlighted in both predicted maps and the annotation.
    }
    
    \label{figure:hero}

\end{figure}

\section{Introduction}

% flow: GEN AI is becoming better
The quality of AI-generated images is improving at a remarkable pace.
Modern text-to-image diffusion models can generate photorealistic images from nothing more than a short text prompt~\cite{Imagen, Glide}.
More recently, novel view synthesis models have been developed that can generate additional viewpoints of a scene starting from a single image~\cite{rombach2021geometry, GenWarp, elata2025novel}, further increasing the realism.
% flow: This is a problem
This rapid progress comes at a serious societal cost, i.e., increasing the risk of fake news, misinformation, and fraud.
% flow: SID exists (different methods)
To address this problem, Synthetic Image Detection (SID) methods have emerged, aiming to distinguish real from generated images.
These methods typically analyze frequency-~\cite{CNN-DCT}, and pixel artifacts~\cite{NPR}, noise residuals~\cite{DnCNN-SID}, pretrained feature vectors~\cite{UnivFD}, and lighting and shadow inconsistencies~\cite{ShadowLines, overview-SID}.

% flow: But it becomes more and more difficult to do single image
A fundamental limitation of these SID approaches is that they analyze an image in isolation, without reasoning about relationships between views of the same scene, i.e., multi-view (MV) image sets.
% flow: However, geometric consistency between views is harder to defeat
In particular, geometric consistency across views is difficult for current generative models to reproduce reliably.
While real photographs of the same scene are inherently consistent because they capture the same physical geometry, generative models are not constrained in this way.
As a result, even if individual views appear plausible, they may geometrically contradict each other, revealing their synthetic origin (e.g., see \Cref{figure:hero}).

Despite the potential of multi-view consistency as a robust signal for synthetic image detection, research in this area remains limited (only a few methods exist~\cite{3D-consistency-scoring, TSED, MEt3R, Zhang2009TwoView, GeCo}) and is primarily focused on evaluating novel view synthesis and video generation models rather than multimedia forensics. 
Existing approaches either produce only global consistency scores, rely on handcrafted geometric constraints, or operate on highly-overlapping image sequences (e.g., adjacent video frames) rather than wide-baseline image pairs (i.e., image pairs with large viewpoint differences and limited overlap). 

In contrast, multimedia forensics requires interpretable pixel-level predictions that localize geometrically inconsistent regions, as well as methods that remain applicable when views differ substantially in viewpoint. 
Additionally, no suitable wide-baseline MV datasets exist that provide controlled geometric inconsistencies together with pixel-level ground-truth annotations, limiting the development and evaluation of localization methods in this setting.

This paper addresses both problems with the following contributions:
(i) We introduce \textbf{DeformView}, a dataset with pixel-level annotations of geometric inconsistencies in wide-baseline multi-view image sets, comprising 1029 objects rendered from 24 viewpoints with both original and deformed variants (\Cref{sec:dataset}).
(ii) We propose \textbf{DEFECt3R}, a lightweight learning-based method for pixel-level geometric inconsistency localization in wide-baseline multi-view image sets. Built upon MEt3R~\cite{MEt3R}, DEFECt3R substantially reduces false positives and improves localization performance (\Cref{sec:method}).
(iii) We establish a benchmark for geometric inconsistency localization in wide-baseline multi-view image sets by evaluating state-of-the-art consistency-scoring methods (TSED~\cite{TSED} and MEt3R~\cite{MEt3R}) alongside DEFECt3R on DeformView, highlighting the limitations of existing methods for multimedia forensics (\Cref{sec:results}).

%-----------------------------%
%         RELATED WORK
%-----------------------------%

% \section{Related Work}
\section{Related Work: Multi-View Consistency}
\label{sec:related_work}

% flow: what is multi-view consistency and why is it a harder-to-defeat signal
Multi-view (MV) consistency refers to the geometric agreement between images of the same scene taken from different viewpoints. In real photographs, this consistency arises naturally, whereas generative models are not inherently constrained to preserve it. This makes MV consistency a powerful signal for detecting synthetic images. Despite its promise, progress in this area is limited by two key challenges: (i) the lack of suitable datasets (\Cref{sec:related_work:datasets}), and (ii) limitations of existing methods (\Cref{sec:related_work:methods}).

\subsection{Multi-View Datasets}
\label{sec:related_work:datasets}

%% flow: Make the statement how there are no dataset available and thus creating and evaluating geometric inconsistencies localization is hard
The first challenge is the lack of suitable datasets.
While recent work has begun exploring geometric inconsistencies, existing datasets do not provide a suitable benchmark for wide-baseline multi-view image pairs.
One related work is by \textbf{Khangaonkar \emph{et al.}}~\cite{object-level-inconsistency-dataset}, who introduce a dataset of 615 image triplets constructed via \textit{object copy-pasting} across views. 
Its goal is to evaluate whether multimodal large language models can identify the geometrically inconsistent object. Therefore, annotations are provided at object level rather than pixel level. 
While these annotations could be converted into segmentation masks, they would reflect the extent of the pasted object rather than the actual geometric inconsistency. 
In particular, they do not distinguish between pixels that violate MV geometry and those that remain plausible, nor do they indicate where the geometric error actually occurs.
As a result, this dataset is not suitable for pixel-level inconsistency localization. 

More recently, Gu \emph{et al.}~\cite{GeCo} introduced \textbf{WarpBench} and \textbf{OccluBench}, two benchmarks for geometric inconsistency localization in generated videos. WarpBench simulates deformations using image-space thin-plate spline warps, while OccluBench focuses on occlusion inconsistency artifacts. 
Although WarpBench provides pixel-level annotations, it applies deformations independently in image space rather than as modifications to a shared 3D geometry.
Consequently, the same deformation cannot be rendered consistently across multiple viewpoints, nor can the benchmark generate geometrically consistent yet visually deformed hard negatives.
Such hard negatives are important when training localization models, as they prevent the model from simply detecting visual deformations instead of genuine cross-view geometric inconsistencies.

\subsection{Multi-View Consistency Methods}
\label{sec:related_work:methods}

% flow: what models exist, and what are their limitations (partially covered in introduction, but now with extra limitations)
The second challenge is the limitations of existing methods.
Early forensic methods relied on handcrafted geometric constraints, such as homographies and fundamental matrices, to detect photographic composites~\cite{Zhang2009TwoView}. 
More recently, only a handful of methods have been proposed to assess geometric consistency in MV image sets, primarily in the context of the evaluation of novel view synthesis and video generation models.
However, these approaches either produce only global consistency scores or operate on highly-overlapping image sequences, restricting their utility for forensic analysis of wide-baseline image pairs.

First, \textbf{3D consistency-scoring}~\cite{3D-consistency-scoring} (3DCS) trains a Neural Radiance Field~\cite{NeRF} on a subset of views and compares rendered novel views to the held-out images.
While effective in principle, this approach requires a large number of views to obtain a reliable reconstruction, and hence is not applicable to only a pair of images. In practice, its performance depends strongly on the quality of the learned NeRF representation, making it less suitable for forensic settings where data is limited and uncontrolled. 

Second, \textbf{TSED}~\cite{TSED} extracts features from image pairs and verifies epipolar consistency given the (required) known camera poses. 
Two images are considered consistent if they have at least $T_{\text{matches}}$ correspondences and the median symmetric epipolar distance of these correspondences is below $T_{\text{error}}$.
TSED's dependence on camera poses complicates deployment in forensic settings, while its notion of consistency is limited to epipolar correspondence agreement rather than explicit reasoning about geometric inconsistencies.

%%% TODO: posible position for scheme

%% This part is a bit bombastic at times
Third, \textbf{MEt3R}~\cite{MEt3R} evaluates MV consistency between image pairs without requiring camera poses. 
It extracts dense DINO~\cite{DINO} feature maps, upsamples them using FeatUp~\cite{FeatUp}, and establishes geometric correspondence using a 3D matching model such as MASt3R~\cite{MASt3R}. Consistency is measured by computing the cosine similarity between aligned features and averaging these into a global score.
Despite improving upon prior methods, MEt3R remains a global metric designed to assess overall agreement between views rather than localize geometric inconsistencies.

More recently, \textbf{GeCo}~\cite{GeCo} proposed dense geometric inconsistency localization for generated videos. GeCo combines optical-flow-based motion consistency with depth-reprojection-based structure consistency to produce interpretable per-pixel inconsistency maps. 
While effective for highly-overlapping video frames, its motion-consistency component relies on dense optical flow and substantial co-visibility between frames, assumptions that become increasingly difficult to satisfy in wide-baseline image pairs. \newline

To address these two challenges, we first introduce a new wide-baseline MV dataset, DeformView, with controlled geometric inconsistencies and pixel-level ground-truth annotations (\Cref{sec:dataset}). 
We then propose DEFECt3R, a learning-based model for geometric inconsistency localization (\Cref{sec:method}) and evaluate its performance on our dataset (\Cref{sec:results}).

%% TODO: original position of model_scheme_5_cropped

%-----------------------------%
%       PROPOSED DATASET
%-----------------------------%

\section{Proposed Dataset: DeformView}
\label{sec:dataset}

%% flow: first say why we need the dataset
To address the lack of suitable datasets for geometric inconsistency localization, we propose \textbf{DeformView}. By applying deformations directly to 3D object meshes, DeformView introduces controlled geometric inconsistencies with known pixel-level ground-truth annotations, while also allowing the same deformation to be rendered consistently from multiple viewpoints (as hard negatives).

%% flow: then how the dataset was formed and why each design choice was made
\textbf{Construction} \titlespace
DeformView is constructed from the Google Scanned Objects collection~\cite{Google_SO}, comprising 1029 3D object meshes. For each object, a localized sinusoidal deformation is applied to a contiguous mesh region covering 8\%--16\% of its faces. The deformation amplitude is sampled uniformly between 4\%--10\% of the object bounding-box diagonal.

Original and deformed meshes are rendered in a shared textured room at a resolution of 512$\times$512\,px. Each object is rendered from 24 viewpoints spaced 15$^\circ$ apart, using a fixed elevation of 20$^\circ$ and a 45$^\circ$ field of view. 
Image pairs are only formed from viewpoints separated by at most 120$^\circ$, resulting in a challenging wide-baseline setting while maintaining sufficient overlap for meaningful cross-view comparison. This yields up to 192 candidate viewpoint pairs per object.

Pixel-level ground-truth masks are obtained by thresholding the per-pixel difference between corresponding renders of the original and deformed meshes. The dataset is split at the object level into training (75\%), validation (15\%), and test (10\%) partitions, ensuring that no object appears in more than one split. This prevents memorization of object-specific geometry or textures and evaluates generalization to previously unseen objects.

Unlike the dataset by Khangaonkar \emph{et al.}~\cite{object-level-inconsistency-dataset} and WarpBench~\cite{GeCo}, which introduce inconsistencies through cross-view copy-pasting and image-space deformations, respectively, DeformView modifies the underlying 3D geometry itself. This enables precise pixel-level annotations, ensures that the same deformation is rendered consistently across multiple viewpoints, and allows for the creation of geometrically consistent yet visually deformed hard negatives.

%% flow: next the pair types and why they are important, and the train/val/test split
\textbf{Pair types}\titlespace
DeformView contains three types of image pairs.
\textbf{Positive pairs} \textit{(original-deformed)} contain one view of the original mesh and one view of the corresponding deformed mesh. 
The resulting geometric inconsistency is confined to the deformed region, and its exact extent is given by the ground-truth mask.
\textbf{Easy negatives} \textit{(original-original)} contain two views of the original mesh and therefore exhibit geometric consistency.
\textbf{Hard negatives} \textit{(deformed-deformed)} contain two views of the same deformed mesh. While the geometry itself is unusual, it remains consistent across viewpoints, requiring methods to verify cross-view disagreement rather than simply identify deformed regions.
Original-deformed and deformed-deformed pairs are only kept if the deformed region is visible in both views

Overall, these design choices yield a dataset with controlled geometric inconsistencies, pixel-level ground-truth annotations, and challenging hard negatives that require reasoning about cross-view consistency rather than appearance alone. DeformView therefore provides a suitable benchmark for evaluating geometric consistency-scoring methods and for training dedicated localization models such as DEFECt3R.

%-----------------------------%
%       PROPOSED METHOD
%-----------------------------%

\section{Proposed Method: DEFECt3R}
\label{sec:method}

%% TODO: move to correct position
\begin{figure*}[tb]
    \centering

    \includegraphics[width=0.94\linewidth, trim=0 4 0 0, clip]{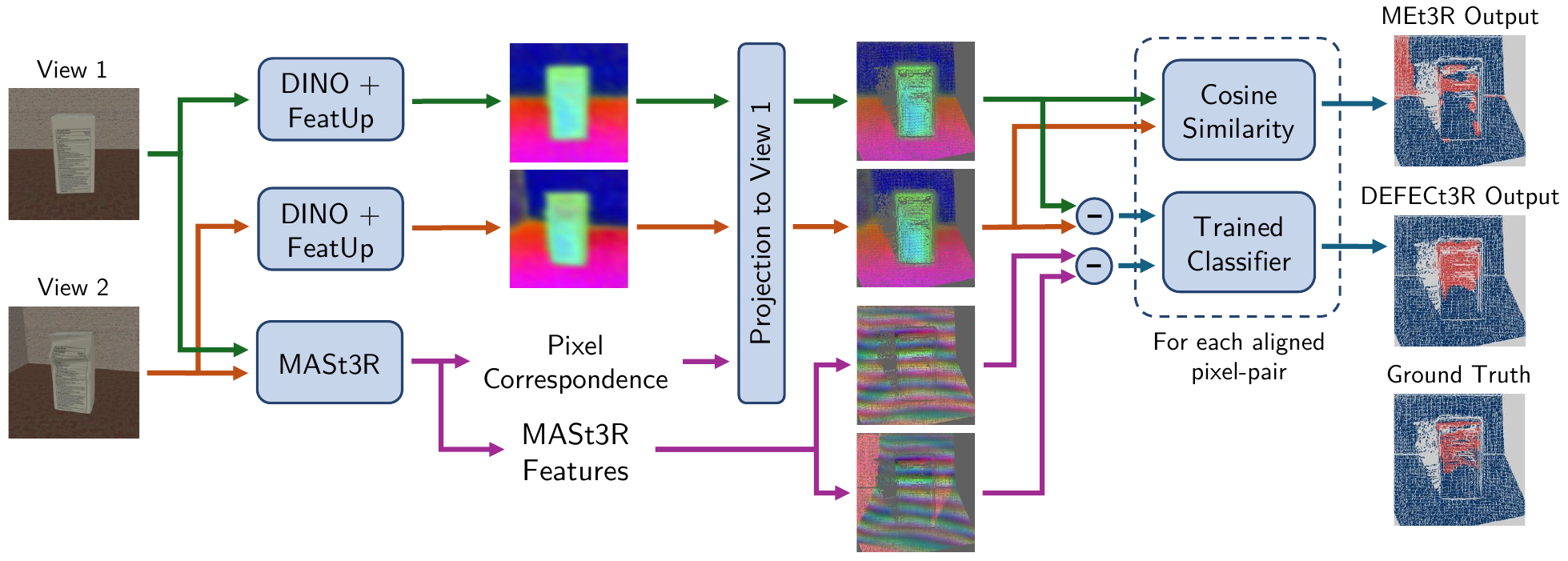} % Drop extension so TAPS can chose PDF for PDF and PNG from HTLM

    \caption{
        Overview of the proposed DEFECt3R method (\Cref{sec:method}) and its relation to MEt3R~\cite{MEt3R}. DINO~\cite{DINO} features are extracted and upsampled using FeatUp~\cite{FeatUp}, then aligned using pixel correspondences estimated by MASt3R~\cite{MASt3R}. The MASt3R pipeline additionally produces feature representations used to establish these correspondences. 
        Whereas MEt3R directly converts cosine similarity between aligned DINO features into an inconsistency score, DEFECt3R uses a lightweight MLP to learn whether DINO and MASt3R feature differences correspond to genuine geometric inconsistencies.
    }
    \Description{
        Pipeline diagram comparing MEt3R and DEFECt3R. Two input images are shown on the left. Both images are processed through DINO feature extraction and MASt3R correspondence estimation. The diagram then splits into two branches. The upper branch represents MEt3R, which computes cosine similarity between aligned DINO features and outputs an inconsistency score. The lower branch represents DEFECt3R, which computes DINO and MASt3R feature differences and feeds them to a multilayer perceptron. The output is a pixel-level inconsistency map.
    }
    \label{fig:defect3r-overview}
\end{figure*}

Geometric inconsistency localization requires determining whether corresponding image regions in different views are mutually consistent. Existing consistency-scoring methods, like MEt3R~\cite{MEt3R}, interpret aligned (DINO) feature dissimilarity as evidence of inconsistency. 
Although geometric inconsistencies naturally lead to differences between corresponding features, feature differences can also arise due to geometrically consistent deformations, or from correspondence errors that cause unrelated image regions to be compared.
Consequently, feature dissimilarity alone is insufficient evidence of geometric inconsistency. 
DEFECt3R addresses this challenge through a supervised per-pixel classifier that learns to reason about cross-view consistency directly from aligned feature pairs. 

\textbf{Design}\titlespace
An overview is shown in \Cref{fig:defect3r-overview}.
DEFECt3R builds on the correspondence pipeline introduced by MEt3R~\cite{MEt3R}. Given an image pair, DINO~\cite{DINO} feature maps are extracted and upsampled using FeatUp~\cite{FeatUp}. Pixel correspondences are then obtained using MASt3R~\cite{MASt3R}, which estimates the camera geometry and internally produces dense feature representations used for aligning (i.e., projecting View 2 to View 1). We use a frozen DINO ViT-S/16 backbone and frozen MASt3R descriptors throughout all experiments.

For each aligned pixel pair, DEFECt3R computes the difference between the aligned DINO features (384-d) as well as the difference between the corresponding MASt3R feature representations (24-d). 
Prior to differencing, all feature vectors are L2-normalized. The resulting feature differences are passed to a lightweight multi-layer perceptron (MLP) classifier.

To prevent the higher-dimensional DINO features from dominating the MASt3R contribution, the DINO difference vector is first projected from 384-d to 64-d dimensions using a two-layer MLP (384→128→64). 
The resulting 64-d embedding is concatenated with the 24-d MASt3R difference vector and processed by a second two-layer MLP (88→64→1) that predicts an inconsistency probability.
Both MLPs use GELU activations and a dropout rate of 0.15.
The complete classifier contains approximately 63k trainable parameters.

%% flow: DINO features provide rich info -> per-pixel classification is possible
Each pixel is classified independently. 
This is possible because both DINO and MASt3R features together already encode rich geometric and contextual information per pixel, allowing the model to reason about cross-view consistency directly from aligned feature pairs. Consequently, no spatial aggregation is required and the model remains lightweight.

\textbf{Training}\titlespace
The classifier is trained on DeformView using focal loss ($\gamma = 4.0$, $\alpha = 0.5$), the AdamW optimizer (LR $= 10^{-3}$, $\lambda = 10^{-3}$), and a cosine annealing learning-rate schedule. Training runs for up to 100 epochs with early stopping based on validation loss.

All foreground pixels are used during training, while background pixels are randomly subsampled to 10\% of the foreground count. This focuses learning on object regions, where inconsistencies occur, while still providing sufficient background examples. 
By training with explicit supervision, including hard negatives from the \textit{deformed-deformed} pair category, the classifier learns to distinguish true geometric inconsistencies from visual deformation.

In summary, DEFECt3R combines aligned cross-view feature differences with supervised learning to distinguish genuine geometric inconsistencies from reprojection artifacts and geometrically consistent deformations, enabling pixel-level localization with substantially fewer false positives than consistency-scoring approaches.

%-----------------------------%
%          RESULTS
%-----------------------------%

\section{Evaluation}
\label{sec:results}

Our proposed dataset and method enable an evaluation of pixel-level geometric inconsistency detection in wide-baseline multi-view image pairs.
We conduct two types of experiments on the DeformView test set:
(i) pixel-level localization, where each pixel is classified and evaluated against the ground truth mask (\Cref{subsec:pixel-level-results}); and
(ii) pair-level classification, where a single consistency decision is made per image pair (\Cref{subsec:pair-level-results}).
We discuss the implications of these results in \Cref{subsec:discussion}.

\textbf{Baseline methods}\titlespace
For pixel-level localization, we compare against MEt3R~\cite{MEt3R}. While MEt3R does not produce pixel-level predictions, its internal cosine similarity map can be interpreted as a per-pixel inconsistency score, providing a natural baseline.

For pair-level classification, we compare against MEt3R~\cite{MEt3R} and TSED~\cite{TSED}. Since TSED requires camera poses, we evaluate it using both ground-truth poses and poses estimated by MASt3R~\cite{MASt3R}.

We do not evaluate 3DCS~\cite{3D-consistency-scoring} or GeCo~\cite{GeCo}, as both are designed for fundamentally different settings. 3DCS requires a sufficiently large set of views to construct a reliable NeRF representation, whereas GeCo operates on highly-overlapping video frames within a temporal sliding window. DeformView instead focuses on wide-baseline image pairs with viewpoint differences of up to 120°.

\input{evaluation-pixel-level-fused-extended.tex}

\textbf{Pair-level scoring}\titlespace
DEFECt3R produces pixel-level inconsistency scores that must be aggregated into a pair-level score. 
Based on validation-set performance, we select the 99th percentile as the aggregation statistic. 
For a fair comparison, MEt3R is evaluated using both its original mean aggregation of the pixel-level similarities and the same 99th-percentile aggregation as DEFECt3R.
TSED requires no aggregation as it directly produces pair-level scores. 

\textbf{Ground-truth correspondence evaluation}\titlespace
To isolate the impact of pose estimation, we additionally evaluate all methods using ground-truth camera poses. 
As correspondences are obtained directly from these poses rather than the MASt3R matching pipeline, the internal MASt3R feature representations are unavailable. Consequently, the ground-truth variant of DEFECt3R operates solely on DINO features.
To separately assess the impact of MASt3R features and correspondence quality, we additionally train a DINO-only variant using MASt3R correspondences.

\textbf{Evaluation metrics}\titlespace
Performance at both the pixel and pair level is evaluated using area under the curve (AUC), F1-score, precision, and recall, capturing both threshold-independent separability and performance at a fixed operating point. 
For pixel-level evaluation, only foreground pixels are considered. This focuses the evaluation on inconsistency localization within the object and avoids bias from the large number of trivially consistent background pixels.
Decision thresholds are selected on the validation set by maximizing the F1 score. For TSED, only the epipolar error threshold $T_{\text{error}}$ is optimized, while $T_{\text{matches}}$ is fixed to its default value of 10. 
For all-negative categories (i.e., \textit{original-original} and \textit{deformed-deformed}), where AUC, F1-score, precision, and recall are undefined, we report the false positive rate (FPR) instead.

\subsection{Pixel-Level Localization}
\label{subsec:pixel-level-results}

\Cref{table:evaluation-pixel-level-fused} summarizes the pixel-level inconsistency localization results on the DeformView test set, both on the entire set and broken down by pair type.
DEFECt3R consistently outperforms MEt3R in terms of AUC, F1, and precision, while substantially reducing false positives. The largest reduction occurs in the deformed regions of the hard-negative \textit{deformed-deformed} pairs, indicating that DEFECt3R is less likely to mistake visually unusual but geometrically consistent regions for inconsistencies.

Comparing DINO-only and DINO+MASt3R variants of DEFECt3R suggests that the feature representations produced by MASt3R provide complementary geometric information beyond the DINO representations alone, resulting in a modest but consistent improvement in localization performance.
The larger gains obtained under the ground-truth pose setting indicate that correspondence quality has an even greater impact on performance.
Ground-truth correspondences affect the two methods differently. For MEt3R, false positive rates decrease substantially, while overall discriminative performance remains largely unchanged. In contrast, DEFECt3R significantly benefits from the improved correspondences, achieving higher AUC, F1-score, and precision.

\input{evaluation-pair-level-split-extended.tex}

While DEFECt3R's recall decreases compared to MEt3R, DEFECt3R achieves substantially higher precision, indicating a more conservative prediction strategy. This suggests that the proposed classifier successfully suppresses many of the erroneous detections produced by the cosine-similarity-based consistency measure, albeit at the cost of missing a larger fraction of true inconsistencies.

\Cref{figure:evaluation-full} provides qualitative examples from all three pair categories.
The consistent \textit{original-original} pairs ((a)--(c)) illustrate a common failure mode of the MASt3R correspondence pipeline. 
Despite the simple scene geometry, large viewpoint differences make correspondence estimation challenging, occasionally resulting in severely incorrect camera poses. 
The resulting reprojection errors cause non-overlapping object regions to be projected onto one another, producing large false-positive inconsistency responses. 
Although driven by the same correspondences, DEFECt3R is able to suppress many of these false positives.

The inconsistent \textit{original-deformed} pairs ((d)--(f)) demonstrate the target scenario of geometric inconsistency localization. While both MEt3R and DEFECt3R respond to the manipulated region, DEFECt3R produces cleaner and more localized predictions with fewer responses in consistent regions.

Finally, rows (g)--(i) show the hard-negative \textit{deformed-deformed} pairs.
Although these regions are geometrically consistent across views, their unusual appearance frequently causes MEt3R to assign high inconsistency scores. In contrast, DEFECt3R largely suppresses these false positives, indicating that it has learned to distinguish visual deformation from genuine cross-view disagreement.

Together, these examples demonstrate how DEFECt3R suppresses false positives caused by correspondence errors and geometrically consistent deformations, while retaining sensitivity to genuine geometric inconsistencies.

\subsection{Pair-Level Classification}
\label{subsec:pair-level-results}

\Cref{table:evaluation-pair-level} reports the pair-level classification results on the DeformView test set.
Both TSED and MEt3R achieve AUCs close to 0.5, indicating little discrimination between consistent and inconsistent image pairs. 
As a result, the F1-optimal operating points correspond to a high-recall regime, yielding recalls close to 1.0 but precisions closely matching the fraction of inconsistent pairs in the test set.

For MEt3R, mean and 99th-percentile aggregation yield very similar results, indicating that aggregation choice has little impact on performance. 
Likewise, while ground-truth correspondences lead to modest improvements for TSED, the method remains close to random performance. Together, these results suggest that neither the aggregation strategy nor correspondence quality fully explains the limitations of existing consistency-scoring methods.

Comparing the DINO-only and full variants of DEFECt3R yields similar conclusions to the pixel-level evaluation. 
While MASt3R features provide a modest improvement, the largest gains are obtained from improved correspondences, resulting in higher AUC, F1, and precision. 
Nevertheless, substantial room for improvement remains even under this idealized correspondence setting.

Overall, DEFECt3R outperforms all baseline methods, demonstrating that learned localization cues provide a more informative basis for pair-level classification than existing consistency scores.

% \input{evaluation-pair-level-split-extended.tex}

% \clearpage

% \subsection{Discussion \& Conclusion}
\section{Discussion \& Conclusion}
\label{subsec:discussion}

%% flow: Existing methods solve a different problem
Most existing consistency-scoring methods were designed for novel view synthesis evaluation rather than multimedia forensics. 
Their objective is to quantify how consistent a set of views is, not to determine whether and where an inconsistency is present. 
As a result, localized inconsistencies that are highly relevant for forensic analysis may only have a minor influence on their global consistency score.

%% flow: Existing methods do not transfer well to the forensic setting
Our experiments on the proposed DeformView dataset support this observation. 
More specifically, the limited impact of alternative aggregation strategies for MEt3R suggests that the primary limitation lies not in its aggregation, but in its underlying cosine-similarity-based consistency signal. 
Similarly, TSED's poor performance, even with ground-truth camera poses, indicates that epipolar correspondences alone are not sufficiently informative for geometric inconsistency detection. 

%% flow: Not only correspondences are a problem
Furthermore, although using ground-truth correspondences instead of MASt3R's estimates substantially reduces MEt3R's false positives, they have little effect on its overall discriminative ability, suggesting that correspondence quality alone does not explain the limitations of existing consistency-scoring approaches.
Nevertheless, qualitative inspection revealed that errors in the correspondences produced by MASt3R remain an important source of false positives.

%% flow: Our method alleviates some of these limitations
DEFECt3R addresses part of these limitations through supervised learning. Rather than interpreting feature differences directly as inconsistencies, it learns to recognize patterns associated with reprojection errors and visually unusual yet geometrically consistent regions.
This substantially reduces false positives and improves localization performance.

The comparison between the DINO-only, DINO+MASt3R, and ground-truth correspondence variants indicates that both feature representation quality and correspondence quality contribute to performance. 
Interestingly, DEFECt3R benefits more from improved (ground-truth) correspondences than MEt3R, suggesting that correspondence quality becomes increasingly important once cross-view feature relationships are interpreted through a learned model rather than directly converted into consistency scores.
Nevertheless, DEFECt3R's precision and recall remain limited, indicating that reliable inconsistency localization remains a challenging open problem for future work.

%% flow: Broader implications + Conclusion
More broadly, our findings suggest that multi-view geometric consistency can serve as a useful foundation for future multimedia forensic methods, provided that correspondence estimation and inconsistency modeling are addressed jointly.
While the consistency-scoring approaches evaluated in this work transfer poorly to forensic analysis, the improvements achieved by DEFECt3R show that learned localization models can exploit information beyond simple consistency metrics and partially compensate for correspondence errors.

A limitation of the current work is that DeformView relies on controlled geometric deformations rather than deformations in AI-generated multi-view imagery. While DeformView's deformations provide precise ground-truth annotations and allow systematic evaluation, the extent to which the observed findings transfer to inconsistencies produced by modern generative models remains an important direction for future research. Extending DeformView with AI-generated multi-view content and evaluating localization methods on such data would provide a valuable next step.

Together, these findings motivate the development of dedicated datasets, benchmarks, and methods for forensic geometric inconsistency localization.
With DeformView and DEFECt3R, we provide a step in this direction.

\input{evaluation-full-M1-M2-M3.tex}

%% file: evaluation-pixel-level-fused-extended.tex
\begin{table*}[tb]
    \centering

    % \multirow{5}{*}{\rotatebox{90}{\parbox{1.25cm}{\centering \shortstack[c]{\text{line 1}\\\text{line 2}}}}}
    % \multirow{3}{*}{\rotatebox{90}{\parbox{1.40cm}{\centering line 1}}}

    \newcommand{\best}[1]{\textbf{#1}}
    \newcommand{\seco}[1]{\underline{#1}}
    \newcommand{\othe}[1]{#1}

    %%% FULL PAGE VERTICAL CENTERING
    % \vspace{1.0em}

    \caption{
        Pixel-level geometric inconsistency localization performance of MEt3R~\cite{MEt3R} and DEFECt3R (\Cref{sec:method}) on the DeformView test set (\Cref{sec:dataset}).
        Results are reported on the full test set and per pair type, using different feature representations (DINO and DINO+MASt3R) and correspondence estimation methods (ground-truth and MASt3R-estimated camera poses).
        Background pixels are excluded to focus the evaluation on inconsistency localization rather than trivial background classification.
        For all-negative categories the FPR is reported instead of AUC, F1-score, precision, and recall.
        Note that the recall is identical for the full test set and the \textit{original-deformed} subset as positive pixels occur exclusively in \textit{original-deformed} pairs.
        Best and second-best results are shown in bold and underlined, respectively.
        DEFECt3R consistently outperforms MEt3R. Both correspondence quality and MASt3R features contribute to performance, with the former having the larger impact.
    }
    \label{table:evaluation-pixel-level-fused}

    \begin{tabular}{l l c c c c c}
        \toprule
    
        \textbf{Subset}
        & \textbf{Metric}    
        % & \textbf{MEt3R}~\cite{MEt3R}
        % & \makecell[c]{
        %     \shortstack[c]{\textbf{MEt3R}\\\cite{MEt3R}}
        % }
        % & \makecell[c]{
        %     \shortstack[c]{\textbf{DEFECt3R}\\\textbf{(ours)}}
        % } \\
        & \multicolumn{2}{c}{\textbf{MEt3R}~\cite{MEt3R}}
        & \multicolumn{3}{c}{\textbf{DEFECt3R (ours)}} \\

        Features
        & & DINO & DINO & DINO+MASt3R & DINO & DINO \\

        Pose estimation
        & & MASt3R & GT & MASt3R & MASt3R & GT \\

        \cmidrule(lr){1-1}
        \cmidrule(lr){2-2}
        \cmidrule(lr){3-4}
        \cmidrule(lr){5-7}

        % \multirow{5}{*}{\rotatebox{90}{\parbox{1.40cm}{\centering Entire Test Set}}}
        \multirow{4}{*}{\centering \textbf{Entire Test Set}}
        % \textbf{Entire Test Set} % just at the top (first row)
        
        & AUC ↑         & \othe{\tr{0.5950}}  & \othe{\tr{0.5950}}  & \seco{\tr{0.7591}}  & \othe{\tr{0.7321}}  & \best{\tr{0.8091}}  \\
        & F1-score ↑    & \othe{\tr{0.0632}}  & \othe{\tr{0.051}}   & \seco{\tr{0.1675}}  & \othe{\tr{0.1451}}  & \best{\tr{0.207}}  \\
        & Precision ↑   & \othe{\tr{0.033}}   & \othe{\tr{0.027}}   & \seco{\tr{0.140}}   & \othe{\tr{0.109}}   & \best{\tr{0.184}}   \\
        & Recall ↑      & \best{\tr{0.640}}   & \seco{\tr{0.390}}   & \othe{\tr{0.208}}   & \othe{\tr{0.218}}   & \othe{\tr{0.238}}   \\

        \dottedline

        % \multirow{5}{*}{\rotatebox{90}{\parbox{1.25cm}{\centering \shortstack[c]{\text{Original}\\\text{-Deform}}}}}
        \multirow{4}{*}{\centering \textbf{Original-Deformed}}
        % \textbf{Original-Deformed} % just at the top (first row)

        & AUC ↑         & \othe{\tr{0.583}}   & \othe{\tr{0.571}}   & \seco{\tr{0.683}}  & \othe{\tr{0.661}}  & \best{\tr{0.723}}  \\
        & F1-score ↑    & \othe{\tr{0.169}}   & \othe{\tr{0.133}}   & \seco{\tr{0.205}}  & \othe{\tr{0.197}}  & \best{\tr{0.235}}  \\
        & Precision ↑   & \othe{\tr{0.098}}   & \othe{\tr{0.080}}   & \seco{\tr{0.202}}  & \othe{\tr{0.180}}  & \best{\tr{0.232}}  \\
        & Recall ↑      & \best{\tr{0.640}}   & \seco{\tr{0.390}}   & \othe{\tr{0.208}}  & \othe{\tr{0.218}}  & \othe{\tr{0.238}}  \\

        \dottedline

        % \multirow{2}{*}{\rotatebox{90}{\parbox{1.25cm}{\centering \shortstack[c]{\text{Deform}\\\text{-Deform}}}}}

        \textbf{Deformed-Deformed} & & & & & \\
        
        \quad deformed region  
        & FPR ↓         & \othe{\tr{0.613}}   & \othe{\tr{0.345}}   & \best{\tr{0.015}}  & \seco{\tr{0.016}}  & \othe{\tr{0.017}}  \\
        \quad non-deformed region  
        & FPR ↓         & \othe{\tr{0.488}}   & \othe{\tr{0.290}}   & \seco{\tr{0.020}}  & \othe{\tr{0.031}}  & \best{\tr{0.009}}  \\
        
        \dottedline

        % \multirow{1}{*}{\rotatebox{90}{\parbox{1.25cm}{\centering \shortstack[c]{\text{Original}\\\text{-Original}}}}}

        \textbf{Original-Original}
        & FPR ↓         & \othe{\tr{0.470}}   & \othe{\tr{0.293}}   & \seco{\tr{0.015}}  & \othe{\tr{0.030}}  & \best{\tr{0.007}}  \\
        
        \bottomrule
    \end{tabular}

    % \vspace{0.5em}
    % \vspace{-0.2em}

\end{table*}

%% file: evaluation-pair-level-split-extended.tex
\begin{table*}[tb]
    \centering
    % \footnotesize
    % \setlength{\tabcolsep}{5.5pt} % column space to make it smaller

    \newcommand{\best}[1]{\textbf{#1}}
    \newcommand{\seco}[1]{\underline{#1}}
    \newcommand{\othe}[1]{#1}

    %%% FULL PAGE VERTICAL CENTERING
    % \vspace{1.5em}

    \caption{
        Pair-level geometric inconsistency classification performance of TSED~\cite{TSED}, MEt3R~\cite{MEt3R}, and DEFECt3R (\Cref{sec:method}) on the DeformView test set (\Cref{sec:dataset}). 
        Results are reported using different feature representations (DINO and DINO+MASt3R), correspondence estimation methods (ground-truth and MASt3R-estimated camera poses), and aggregation strategies (mean and 99th-percentile).
        Best and second-best results are shown in bold and underlined, respectively. 
        Recall values are not highlighted, as they primarily reflect the selected operating point rather than the underlying discriminative ability of the model.
        DEFECt3R substantially outperforms existing consistency-scoring methods. Similar to the pixel-level results, the largest gains are obtained from improved correspondences, while MASt3R features provide an additional benefit. 
    }
    \label{table:evaluation-pair-level}

    \begin{tabular}{l c c c c c c c c c}
        
        \toprule
        
        % Subset & 
        \textbf{Metric}
        & \multicolumn{2}{c}{\textbf{TSED}~\cite{TSED}}
        
        & \multicolumn{4}{c}{\textbf{MEt3R}~\cite{MEt3R}}
        
        & \multicolumn{3}{c}{\textbf{DEFECt3R (ours)}} \\

        Features
        & & & DINO & DINO & DINO & DINO & DINO+MASt3R & DINO & DINO \\
        
        Aggregation
        % & &  
        % & \multicolumn{2}{c}{Mean} 
        % & \multicolumn{2}{c}{99th} 
        % & \multicolumn{3}{c}{99th}  \\
        & & & Mean & Mean & 99th & 99th & 99th & 99th & 99th \\

        Pose estimation
        & MASt3R & GT & MASt3R & GT & MASt3R & GT & MASt3R & MASt3R & GT \\

        % & Known & MASt3R & Mean & 99.7th & 99.7th \\
        
        % \cmidrule(lr){1-1}
        % \cmidrule(lr){2-2}
        % \cmidrule(lr){3-5}
        
        \cmidrule(lr){1-1}
        \cmidrule(lr){2-3}
        % \cmidrule(lr){4-5}
        % \cmidrule(lr){6-7}
        \cmidrule(lr){4-7}
        \cmidrule(lr){8-10}

        AUC ↑            & \othe{\tr{0.4835}} & \othe{\tr{0.5204}} & \othe{\tr{0.4997}} & \othe{\tr{0.5155}} & \othe{\tr{0.5130}} & \othe{\tr{0.5157}} & \seco{\tr{0.7126}} & \othe{\tr{0.6788}} & \best{\tr{0.7373}} \\
        % &
        F1-score ↑       & \othe{\tr{0.3159}} & \othe{\tr{0.3253}} & \othe{\tr{0.4822}} & \othe{\tr{0.4803}} & \othe{\tr{0.4810}} & \othe{\tr{0.4776}} & \seco{\tr{0.5370}} & \othe{\tr{0.5282}} & \best{\tr{0.5579}}  \\
        % & 
        Precision ↑      & \othe{\tr{0.3159}} & \othe{\tr{0.3253}} & \othe{\tr{0.319}}  & \othe{\tr{0.317}}  & \othe{\tr{0.319}}  & \othe{\tr{0.314}}  & \othe{\tr{0.394}}  & \seco{\tr{0.404}}  & \best{\tr{0.510}}  \\
        % & 
        % Recall ↑         & \othe{\tr{0.9042}} & \othe{\tr{0.9583}} & \seco{\tr{0.992}}  & \best{\tr{0.993}}  & \othe{\tr{0.977}}  & \best{\tr{0.993}}  & \othe{\tr{0.843}}  & \othe{\tr{0.764}}  & \othe{\tr{0.616}}  \\
        % & NON HIGHTLIGHTED
        Recall ↑         & \tr{0.9042} & \tr{0.9583} & \tr{0.992}  & \tr{0.993}  & \tr{0.977}  & \tr{0.993}  & \tr{0.843}  & \tr{0.764}  & \tr{0.616}  \\

        \bottomrule
        
    \end{tabular}

    % %%% FULL PAGE VERTICAL CENTERING
    % \vspace{6em}
    % \vspace{-0.5em}
    
\end{table*}

%% file: evaluation-full-M1-M2-M3.tex
\begin{figure*}[p]

    % \newcommand{\verticalcolumn}{0.025\linewidth}
    % \newcommand{\subfigwidth}{0.100\linewidth} % Define a variable for the width
    % % space inbetween = 1 - 0.025 - 9 * 0.100 = 0.075
    % % space between 2 cols = 0.075 / 10 = 0.0075
    % \newcommand{\figureheight}{0.100\linewidth}
    % % width of 2 wide = 2 * 0.100 + 1 * 0.0060 = 0.2075
    % \newcommand{\subfigwidthTWO}{0.2075\linewidth}
    % % width of 4 wide = 4 * 0.100 + 3 * 0.0060 = 0.4225
    % \newcommand{\subfigwidthFOUR}{0.4225\linewidth}
    % % width of 3 wide = 3 * 0.100 + 2 * 0.0060 = 0.315
    % \newcommand{\subfigwidthTHREE}{0.315\linewidth}

    % % \newcommand{\collablesmall}[1]{\textbf{#1}} 

    % \newcommand{\imagespacing}{0.3em}

    \centering

    %%% FULL PAGE VERTICAL CENTERING
    \vspace{1.0em}

    % Column labels
    \begin{minipage}[c]{0.025\linewidth}
        \phantom{\textbf{(a)}}
    \end{minipage}%
    \hfill
    \begin{minipage}[c]{0.2075\linewidth}
        \centering
        \phantom{\textbf{Image 1}}
    \end{minipage}
    \hfill
    \begin{minipage}[c]{0.4225\linewidth}
        \centering
        \textbf{MASt3R Pose-Estimation}
    \end{minipage}
    \hfill
    \begin{minipage}[c]{0.315\linewidth}
        \centering
        \textbf{GT Poses}
    \end{minipage}

    \vspace{0.8em} % Add vertical spacing between labels and images

    % --------------------------------
    %           LABLES
    % --------------------------------

    % Column labels
    \begin{minipage}[c]{0.025\linewidth}
        \phantom{\textbf{(a)}}
    \end{minipage}%
    \hfill
    \begin{minipage}[c]{0.100\linewidth}
        \centering
        \textbf{View 1}
    \end{minipage}
    \hfill
    \begin{minipage}[c]{0.100\linewidth}
        \centering
        \textbf{View 2}
    \end{minipage}
    \hfill
    % ========================================================================================================================================
    \begin{minipage}[c]{0.100\linewidth}
        \centering
        \shortstack[c]{\textbf{Ground}\\\textbf{Truth}}
    \end{minipage}
    \hfill
    \begin{minipage}[c]{0.100\linewidth}
        \centering
        % \textbf{MEt3R pred.}
        \shortstack[c]{\textbf{MEt3R}\\\textbf{pred.}}
    \end{minipage}
    \hfill
    \begin{minipage}[c]{0.100\linewidth}
        \centering
        % \textbf{MEt3R pred.}
        \shortstack[c]{\textbf{DEFECt3R}\\\textbf{pred. (ours)}}
    \end{minipage}
    \hfill
    \begin{minipage}[c]{0.100\linewidth}
        \centering
        \shortstack[c]{\textbf{DEFECt3R}\\\textbf{pred. (ours)}}
    \end{minipage}
    \hfill
    % ========================================================================================================================================
    \begin{minipage}[c]{0.100\linewidth}
        \centering
        \shortstack[c]{\textbf{Ground}\\\textbf{Truth}}
    \end{minipage}
    \hfill
    \begin{minipage}[c]{0.100\linewidth}
        \centering
        % \textbf{MEt3R pred.}
        \shortstack[c]{\textbf{MEt3R}\\\textbf{pred.}}
    \end{minipage}
    \hfill
    \begin{minipage}[c]{0.100\linewidth}
        \centering
        \shortstack[c]{\textbf{DEFECt3R}\\\textbf{pred. (ours)}}
    \end{minipage}

    \vspace{0.2em} % Add vertical spacing between labels and images

    % --------------------------------
    %           FEATURES
    % --------------------------------

    % Column labels
    \begin{minipage}[c]{0.025\linewidth}
        \phantom{\textbf{(a)}}
    \end{minipage}%
    \hfill
    \begin{minipage}[c]{0.100\linewidth}
        \centering
        \phantom{\textbf{Image 1}}
    \end{minipage}
    \hfill
    \begin{minipage}[c]{0.100\linewidth}
        \centering
        \phantom{\textbf{Image 1}}
    \end{minipage}
    \hfill
    % ========================================================================================================================================
    \begin{minipage}[c]{0.100\linewidth}
        \centering
        \phantom{\textbf{Image 1}}
    \end{minipage}
    \hfill
    \begin{minipage}[c]{0.100\linewidth}
        \centering
        % \textbf{MEt3R pred.}
        \textbf{DINO}
    \end{minipage}
    \hfill
    \begin{minipage}[c]{0.100\linewidth}
        \centering
        % \textbf{MEt3R pred.}
        \textbf{DINO}
    \end{minipage}
    \hfill
    \begin{minipage}[c]{0.100\linewidth}
        \centering
        \shortstack[c]{\textbf{DINO +}\\\textbf{MASt3R}}
    \end{minipage}
    \hfill
    % ========================================================================================================================================
    \begin{minipage}[c]{0.100\linewidth}
        \centering
        \phantom{\textbf{Image 1}}
    \end{minipage}
    \hfill
    \begin{minipage}[c]{0.100\linewidth}
        \centering
        % \textbf{MEt3R pred.}
        \textbf{DINO}
    \end{minipage}
    \hfill
    \begin{minipage}[c]{0.100\linewidth}
        \centering
        \textbf{DINO}
    \end{minipage}

    \vspace{0.3em} % Add vertical spacing between labels and images

    % -----------------------------------------
    %        ROW 3 (REAL-REAl) (pair_00400)
    % -----------------------------------------
    
    {
        \begin{minipage}[c]{0.025\linewidth}
            \textbf{(a)}
        \end{minipage}%
        \hfill
        \begin{subfigure}[c]{0.100\linewidth}
            \centering
            \includegraphics[width=\linewidth]{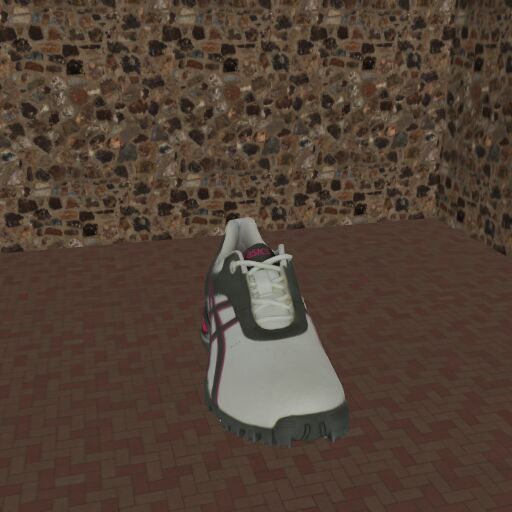}
        \end{subfigure}
        \hfill
        \begin{subfigure}[c]{0.100\linewidth}
            \centering
            \includegraphics[width=\linewidth]{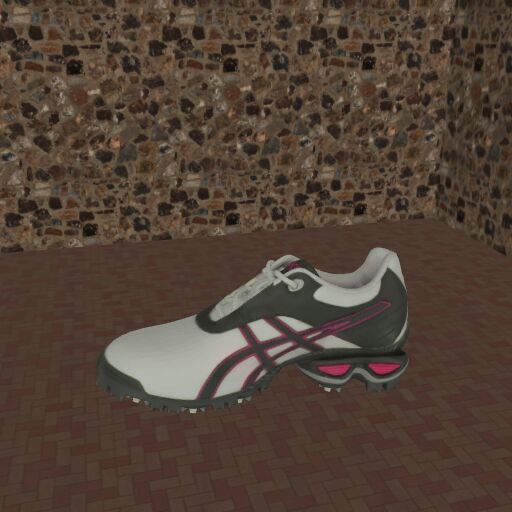}
        \end{subfigure}
        \hfill
        % ========================================================================================================================================
        \begin{subfigure}[c]{0.100\linewidth}
            \centering
            \includegraphics[width=\linewidth]{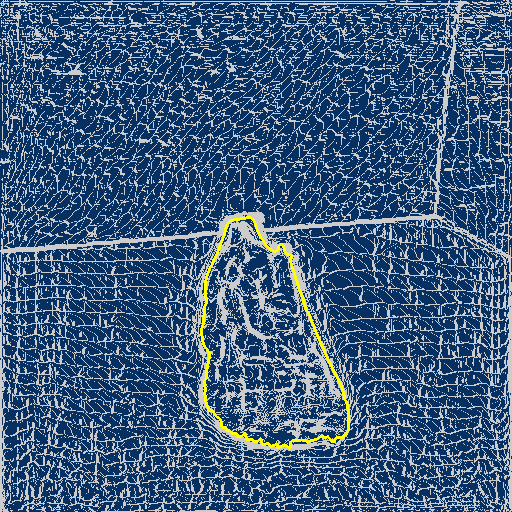}
        \end{subfigure}
        \hfill
        \begin{subfigure}[c]{0.100\linewidth}
            \centering
            \includegraphics[width=\linewidth]{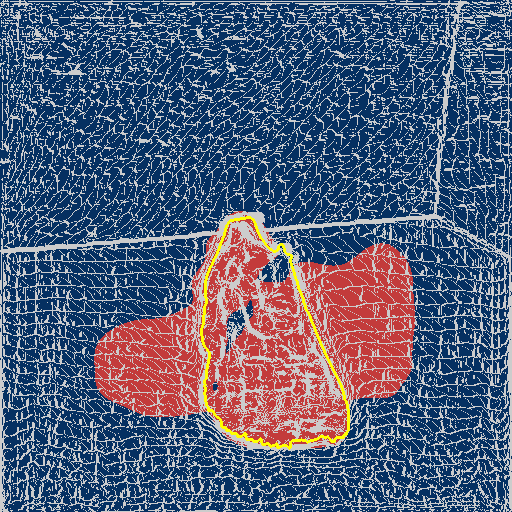}
        \end{subfigure}
        \hfill
        \begin{subfigure}[c]{0.100\linewidth}
            \centering
            \includegraphics[width=\linewidth]{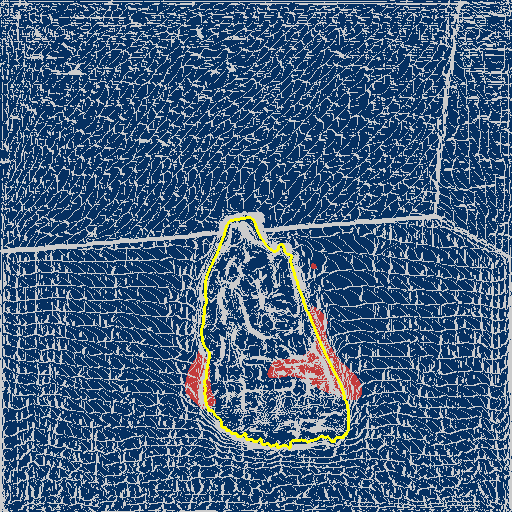}
        \end{subfigure}
        \hfill
        \begin{subfigure}[c]{0.100\linewidth}
            \centering
            \includegraphics[width=\linewidth]{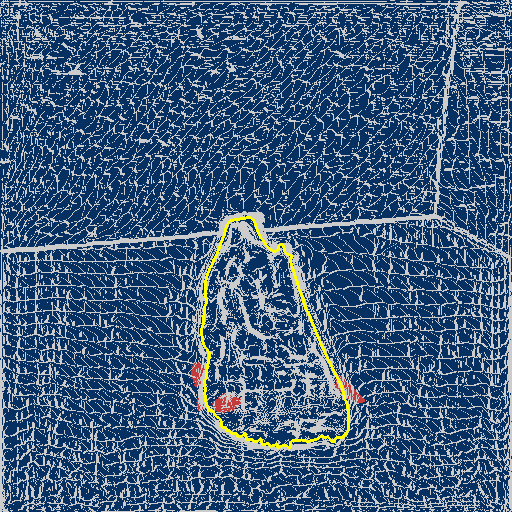}
        \end{subfigure}
        \hfill
        % ========================================================================================================================================
        \begin{subfigure}[c]{0.100\linewidth}
            \centering
            \includegraphics[width=\linewidth]{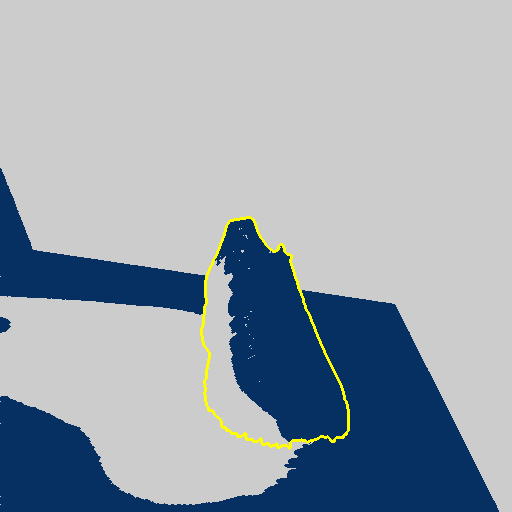}
        \end{subfigure}
        \hfill
        \begin{subfigure}[c]{0.100\linewidth}
            \centering
            \includegraphics[width=\linewidth]{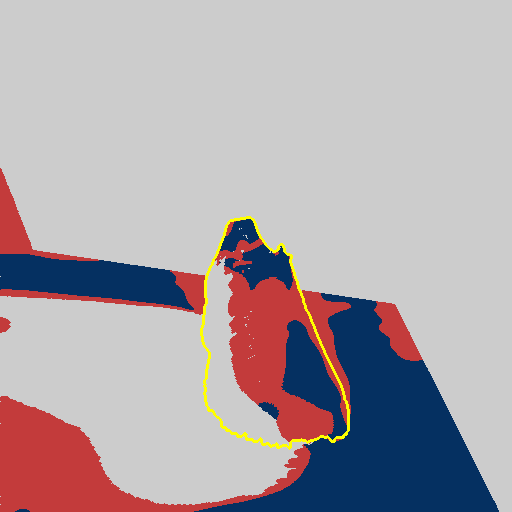}
        \end{subfigure}
        \hfill
        \begin{subfigure}[c]{0.100\linewidth}
            \centering
            \includegraphics[width=\linewidth]{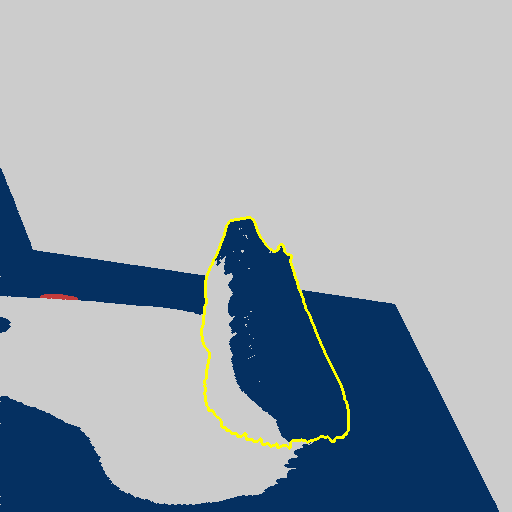}
        \end{subfigure}

        \vspace{0.3em} % Add vertical spacing between labels and images
    }

    % -----------------------------------------
    %        ROW 4 (REAL-REAl) (pair_01516)
    % -----------------------------------------
    
    {
        \begin{minipage}[c]{0.025\linewidth}
            \textbf{(b)}
        \end{minipage}%
        \hfill
        \begin{subfigure}[c]{0.100\linewidth}
            \centering
            \includegraphics[width=\linewidth]{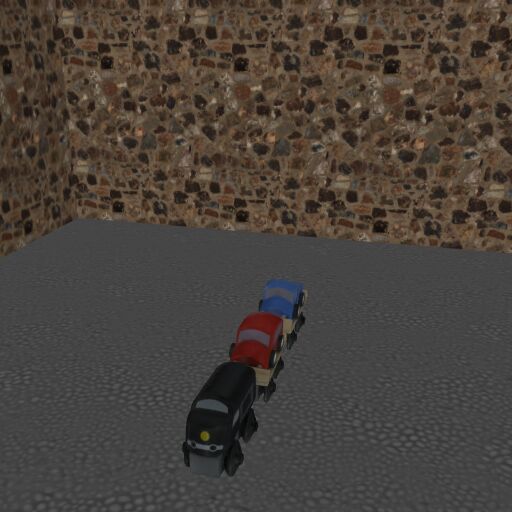}
        \end{subfigure}
        \hfill
        \begin{subfigure}[c]{0.100\linewidth}
            \centering
            \includegraphics[width=\linewidth]{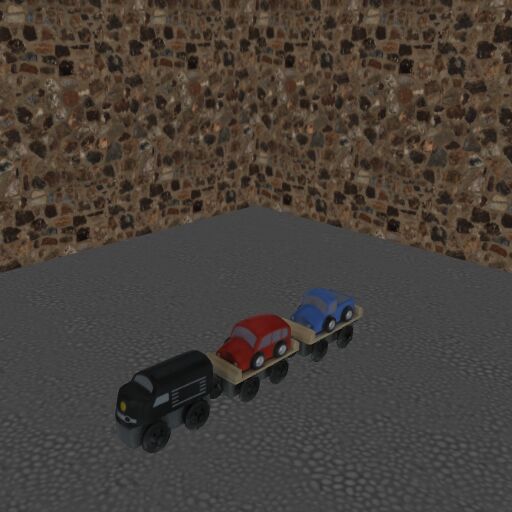}
        \end{subfigure}
        \hfill
        % ========================================================================================================================================
        \begin{subfigure}[c]{0.100\linewidth}
            \centering
            \includegraphics[width=\linewidth]{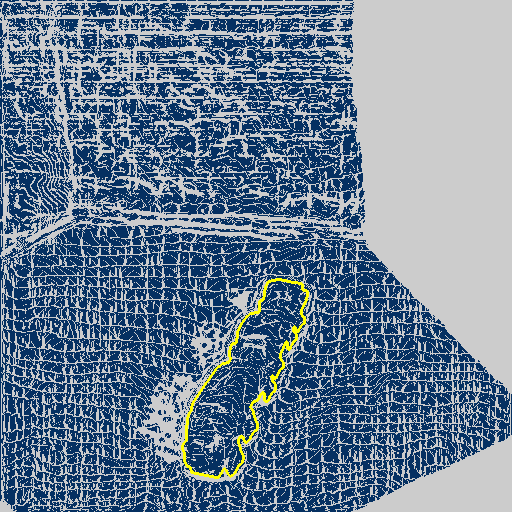}
        \end{subfigure}
        \hfill
        \begin{subfigure}[c]{0.100\linewidth}
            \centering
            \includegraphics[width=\linewidth]{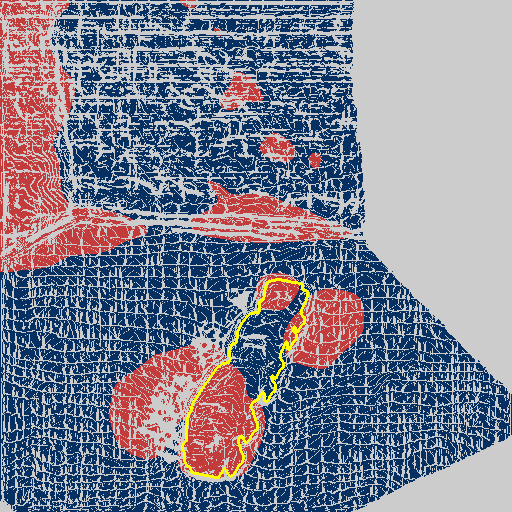}
        \end{subfigure}
        \hfill
        \begin{subfigure}[c]{0.100\linewidth}
            \centering
            \includegraphics[width=\linewidth]{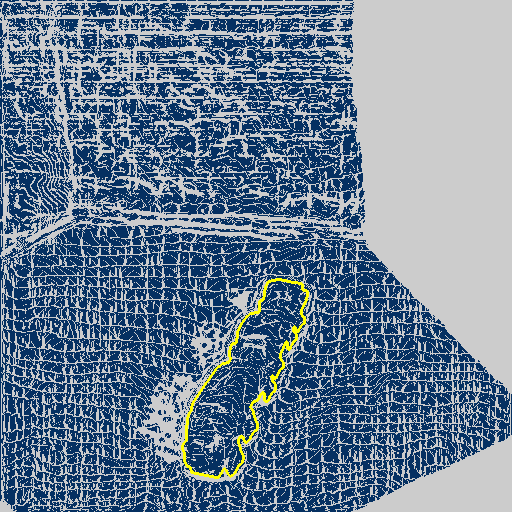}
        \end{subfigure}
        \hfill
        \begin{subfigure}[c]{0.100\linewidth}
            \centering
            \includegraphics[width=\linewidth]{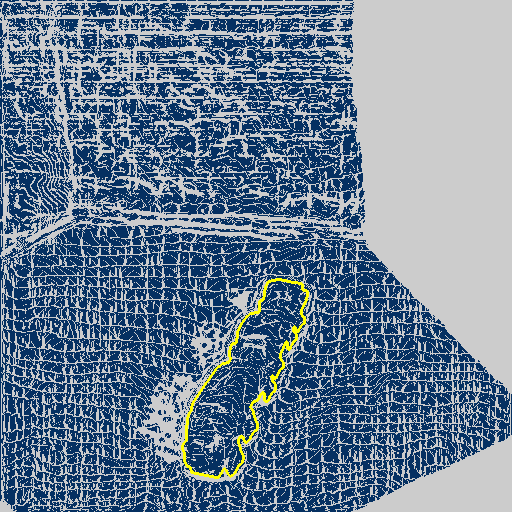}
        \end{subfigure}
        \hfill
        % ========================================================================================================================================
        \begin{subfigure}[c]{0.100\linewidth}
            \centering
            \includegraphics[width=\linewidth]{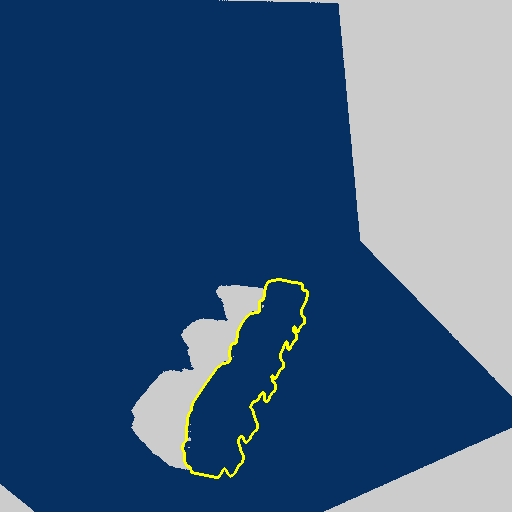}
        \end{subfigure}
        \hfill
        \begin{subfigure}[c]{0.100\linewidth}
            \centering
            \includegraphics[width=\linewidth]{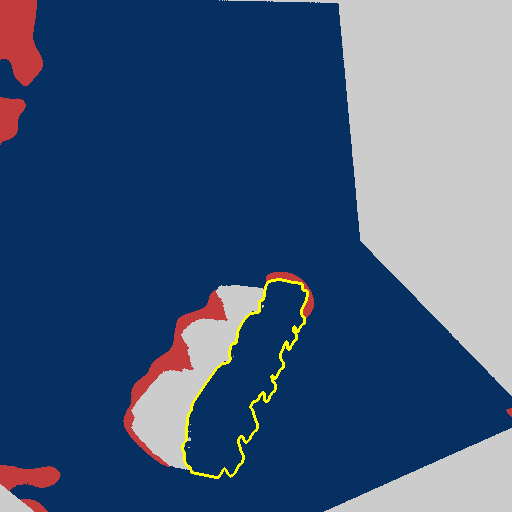}
        \end{subfigure}
        \hfill
        \begin{subfigure}[c]{0.100\linewidth}
            \centering
            \includegraphics[width=\linewidth]{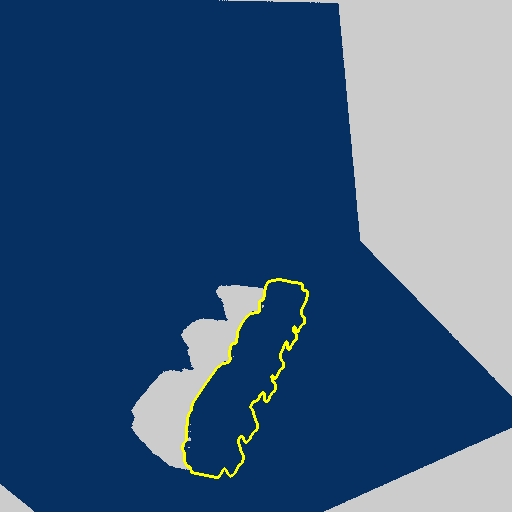}
        \end{subfigure}

        \vspace{0.3em} % Add vertical spacing between labels and images
    }

    % -----------------------------------------
    %        ROW 5 (REAL-REAl) (pair_04360)
    % -----------------------------------------

    {
        \begin{minipage}[c]{0.025\linewidth}
            \textbf{(c)}
        \end{minipage}%
        \hfill
        \begin{subfigure}[c]{0.100\linewidth}
            \centering
            \includegraphics[width=\linewidth]{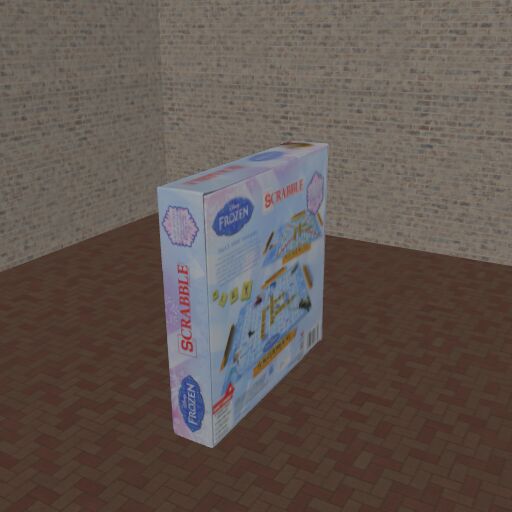}
        \end{subfigure}
        \hfill
        \begin{subfigure}[c]{0.100\linewidth}
            \centering
            \includegraphics[width=\linewidth]{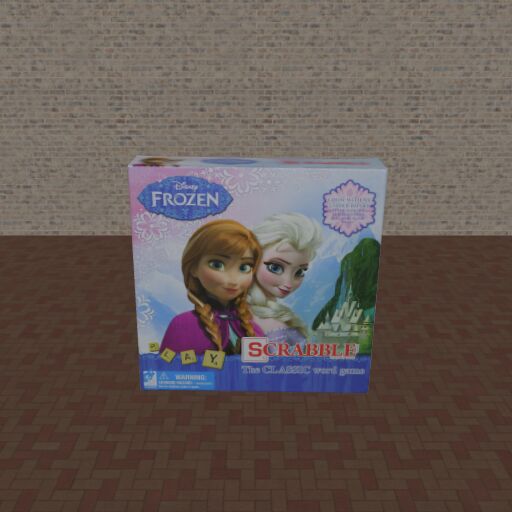}
        \end{subfigure}
        \hfill
        % ========================================================================================================================================
        \begin{subfigure}[c]{0.100\linewidth}
            \centering
            \includegraphics[width=\linewidth]{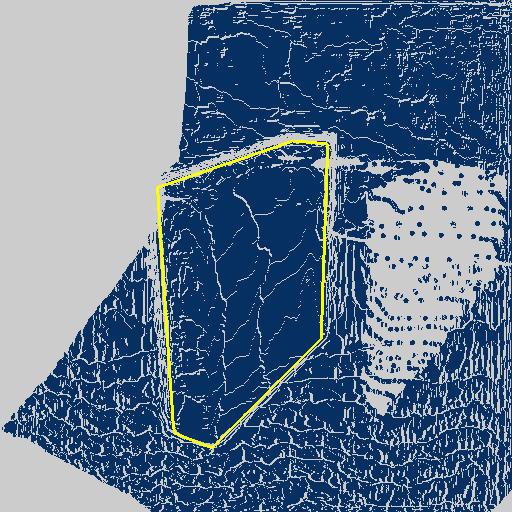}
        \end{subfigure}
        \hfill
        \begin{subfigure}[c]{0.100\linewidth}
            \centering
            \includegraphics[width=\linewidth]{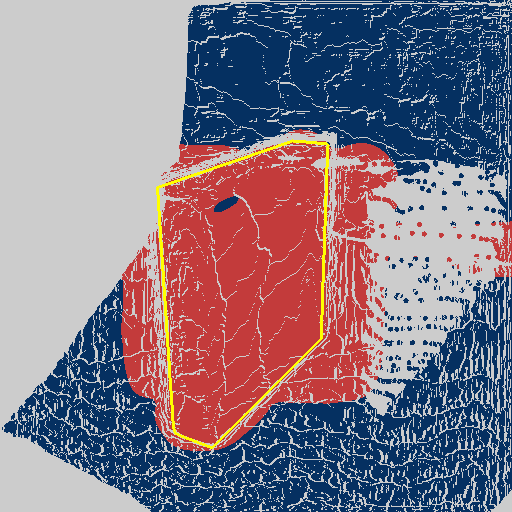}
        \end{subfigure}
        \hfill
        \begin{subfigure}[c]{0.100\linewidth}
            \centering
            \includegraphics[width=\linewidth]{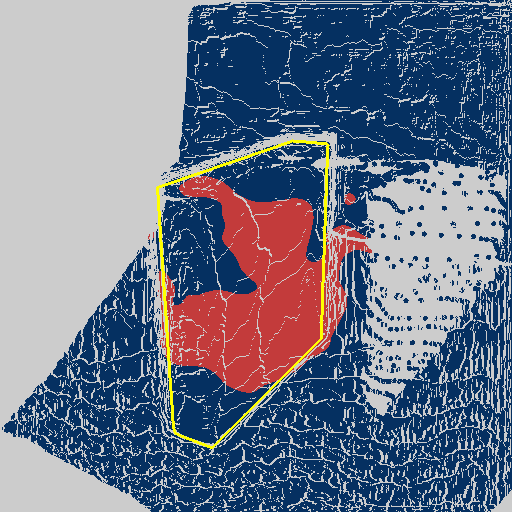}
        \end{subfigure}
        \hfill
        \begin{subfigure}[c]{0.100\linewidth}
            \centering
            \includegraphics[width=\linewidth]{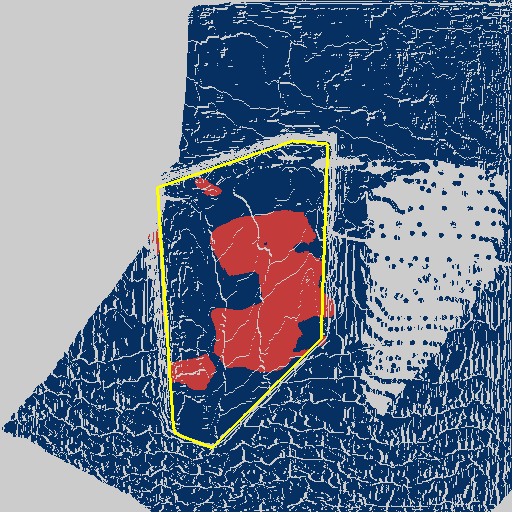}
        \end{subfigure}
        \hfill
        % ========================================================================================================================================
        \begin{subfigure}[c]{0.100\linewidth}
            \centering
            \includegraphics[width=\linewidth]{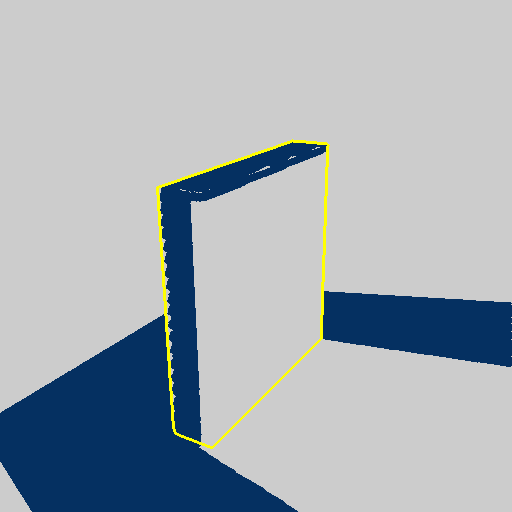}
        \end{subfigure}
        \hfill
        \begin{subfigure}[c]{0.100\linewidth}
            \centering
            \includegraphics[width=\linewidth]{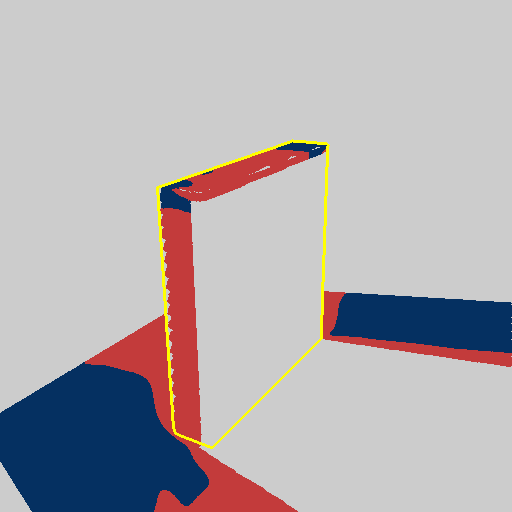}
        \end{subfigure}
        \hfill
        \begin{subfigure}[c]{0.100\linewidth}
            \centering
            \includegraphics[width=\linewidth]{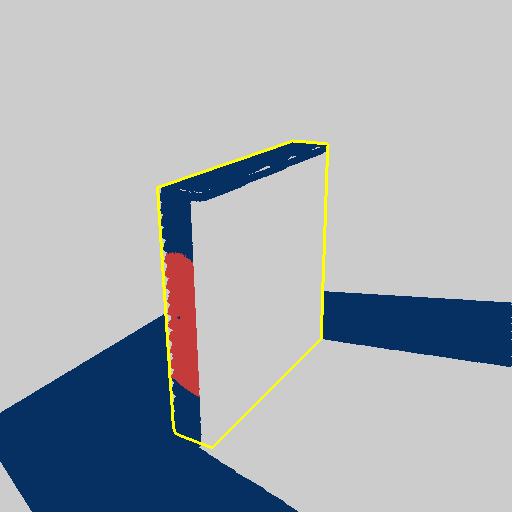}
        \end{subfigure}

        \vspace{0.3em} % Add vertical spacing between labels and images
    }

    % -----------------------------------------
    %        ROW 6 (REAL-DEF) (pair_00645)
    % -----------------------------------------
    
    {
        \begin{minipage}[c]{0.025\linewidth}
            \textbf{(d)}
        \end{minipage}%
        \hfill
        \begin{subfigure}[c]{0.100\linewidth}
            \centering
            \includegraphics[width=\linewidth]{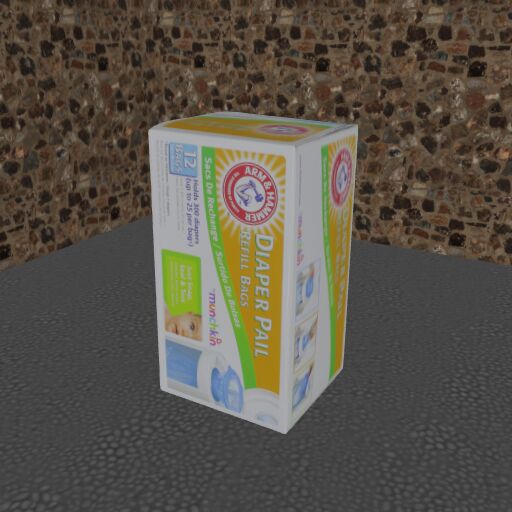}
        \end{subfigure}
        \hfill
        \begin{subfigure}[c]{0.100\linewidth}
            \centering
            \includegraphics[width=\linewidth]{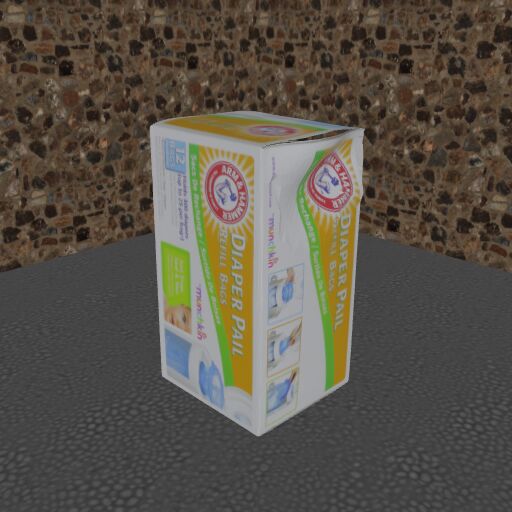}
        \end{subfigure}
        \hfill
        % ========================================================================================================================================
        \begin{subfigure}[c]{0.100\linewidth}
            \centering
            \includegraphics[width=\linewidth]{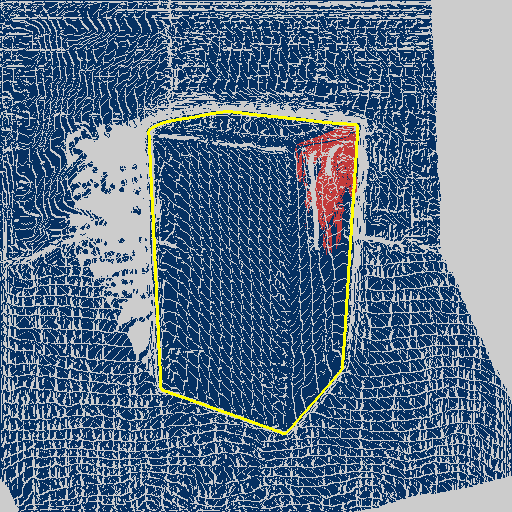}
        \end{subfigure}
        \hfill
        \begin{subfigure}[c]{0.100\linewidth}
            \centering
            \includegraphics[width=\linewidth]{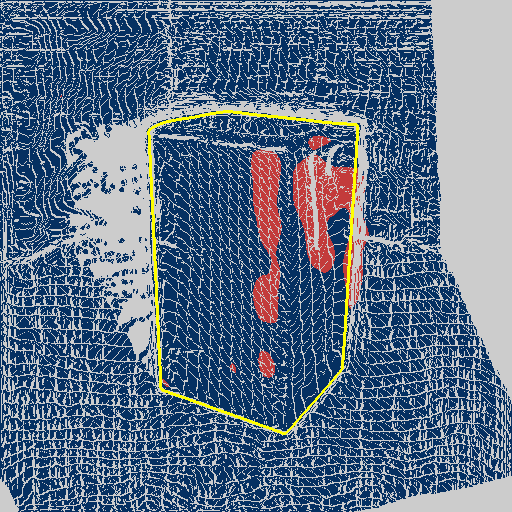}
        \end{subfigure}
        \hfill
        \begin{subfigure}[c]{0.100\linewidth}
            \centering
            \includegraphics[width=\linewidth]{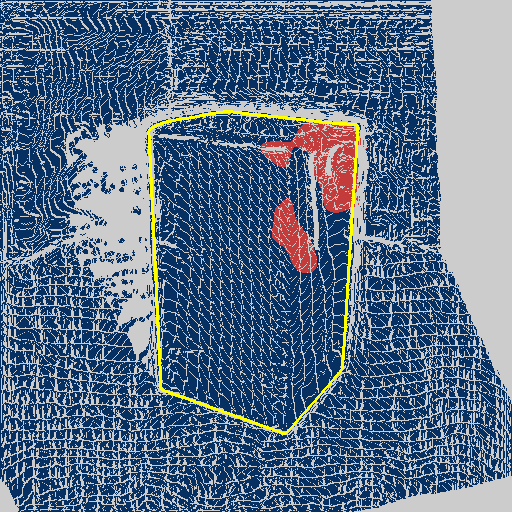}
        \end{subfigure}
        \hfill
        \begin{subfigure}[c]{0.100\linewidth}
            \centering
            \includegraphics[width=\linewidth]{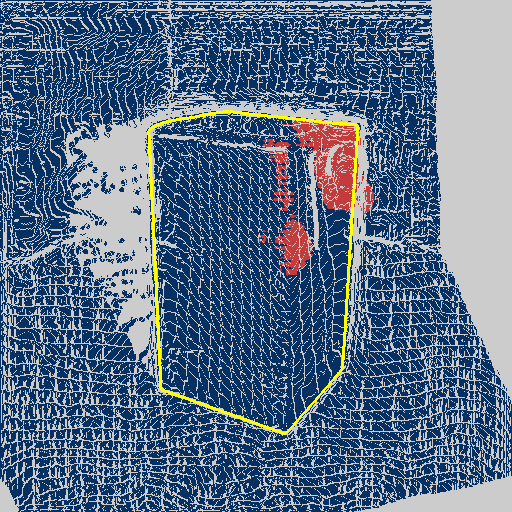}
        \end{subfigure}
        \hfill
        % ========================================================================================================================================
        \begin{subfigure}[c]{0.100\linewidth}
            \centering
            \includegraphics[width=\linewidth]{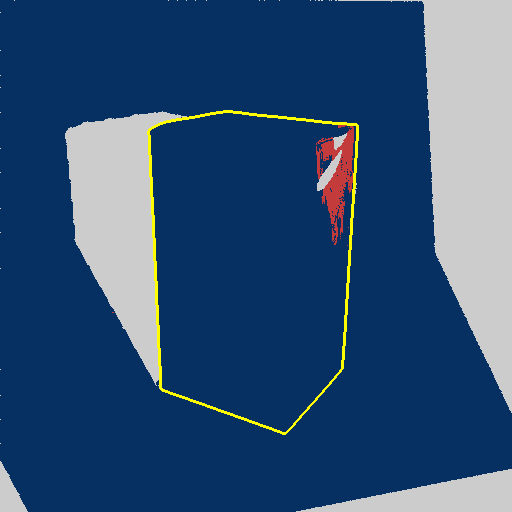}
        \end{subfigure}
        \hfill
        \begin{subfigure}[c]{0.100\linewidth}
            \centering
            \includegraphics[width=\linewidth]{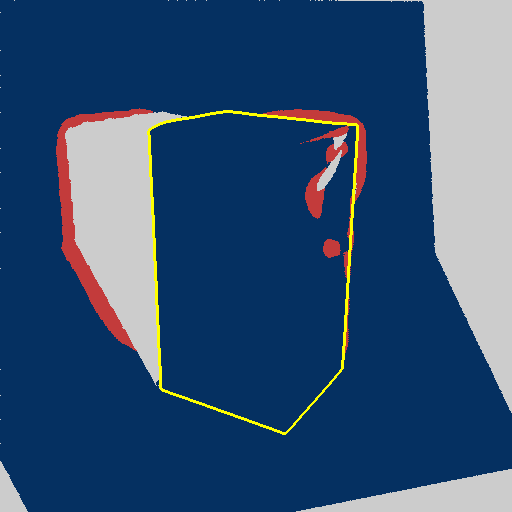}
        \end{subfigure}
        \hfill
        \begin{subfigure}[c]{0.100\linewidth}
            \centering
            \includegraphics[width=\linewidth]{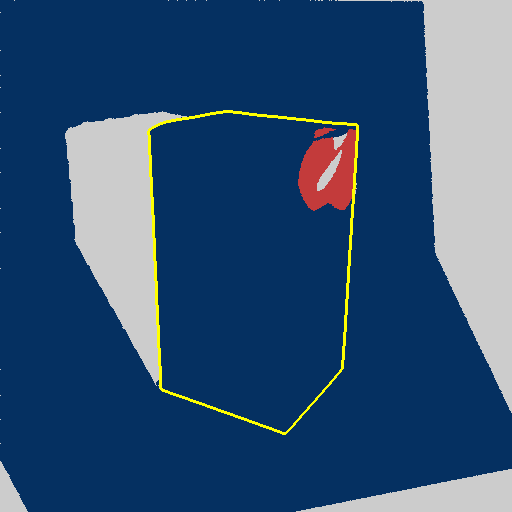}
        \end{subfigure}

        \vspace{0.3em} % Add vertical spacing between labels and images
    }

    % -----------------------------------------
    %        ROW 8 (REAL-DEF) (pair_12132)
    % -----------------------------------------
    
    {
        \begin{minipage}[c]{0.025\linewidth}
            \textbf{(e)}
        \end{minipage}%
        \hfill
        \begin{subfigure}[c]{0.100\linewidth}
            \centering
            \includegraphics[width=\linewidth]{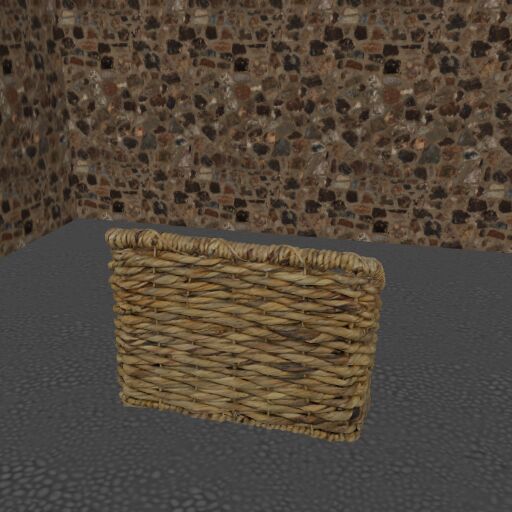}
        \end{subfigure}
        \hfill
        \begin{subfigure}[c]{0.100\linewidth}
            \centering
            \includegraphics[width=\linewidth]{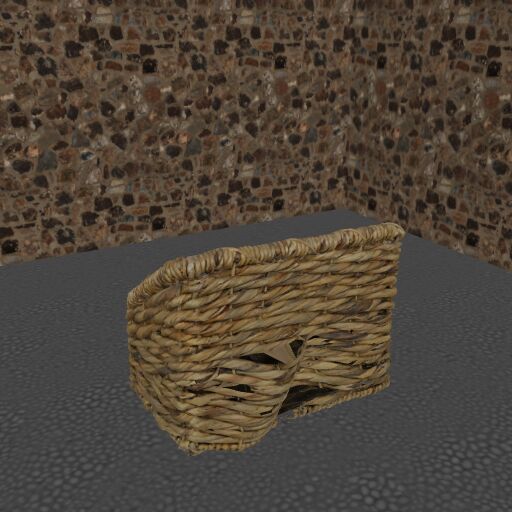}
        \end{subfigure}
        \hfill
        % ========================================================================================================================================
        \begin{subfigure}[c]{0.100\linewidth}
            \centering
            \includegraphics[width=\linewidth]{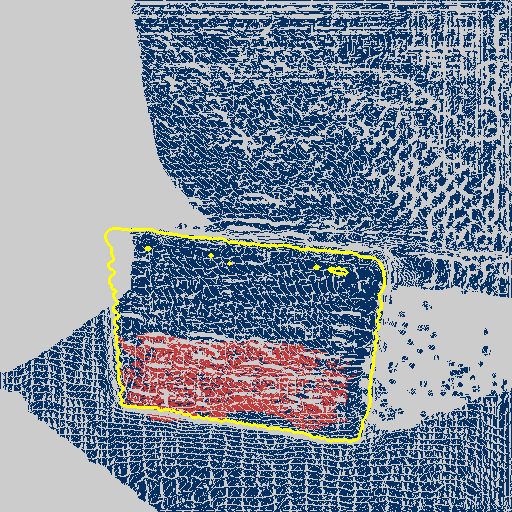}
        \end{subfigure}
        \hfill
        \begin{subfigure}[c]{0.100\linewidth}
            \centering
            \includegraphics[width=\linewidth]{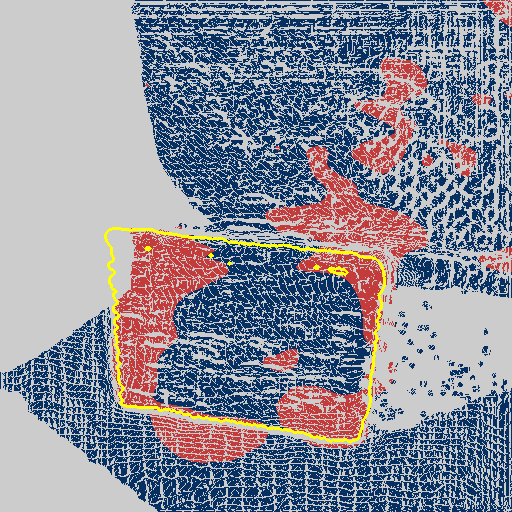}
        \end{subfigure}
        \hfill
        \begin{subfigure}[c]{0.100\linewidth}
            \centering
            \includegraphics[width=\linewidth]{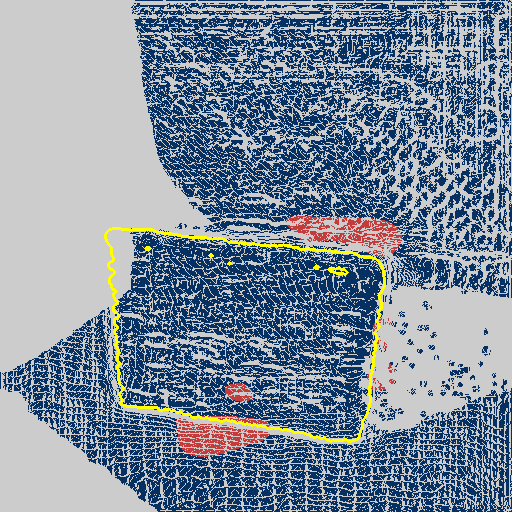}
        \end{subfigure}
        \hfill
        \begin{subfigure}[c]{0.100\linewidth}
            \centering
            \includegraphics[width=\linewidth]{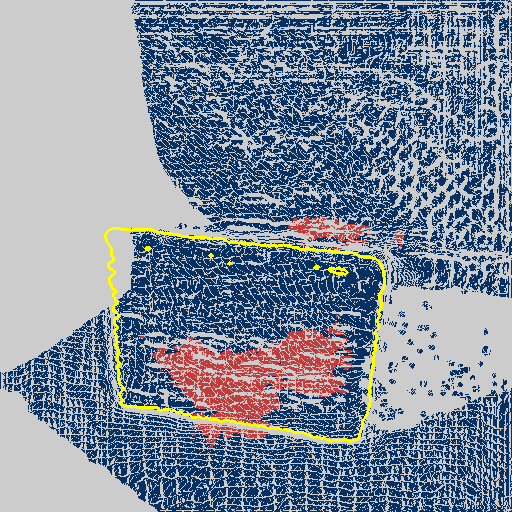}
        \end{subfigure}
        \hfill
        % ========================================================================================================================================
        \begin{subfigure}[c]{0.100\linewidth}
            \centering
            \includegraphics[width=\linewidth]{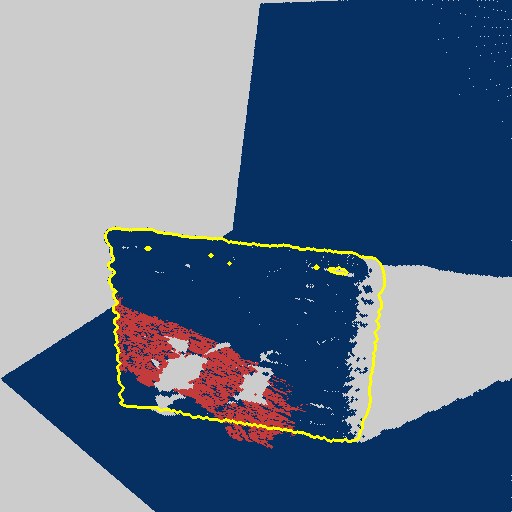}
        \end{subfigure}
        \hfill
        \begin{subfigure}[c]{0.100\linewidth}
            \centering
            \includegraphics[width=\linewidth]{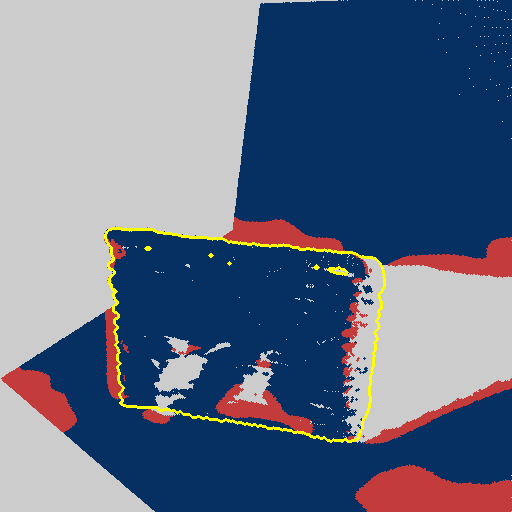}
        \end{subfigure}
        \hfill
        \begin{subfigure}[c]{0.100\linewidth}
            \centering
            \includegraphics[width=\linewidth]{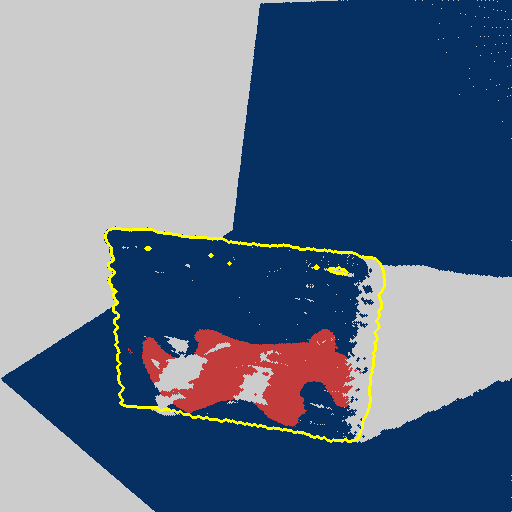}
        \end{subfigure}

        \vspace{0.3em} % Add vertical spacing between labels and images
    }

    % -----------------------------------------
    %        ROW 11 (REAL-DEF) (pair_12745)
    % -----------------------------------------
    
    {
        \begin{minipage}[c]{0.025\linewidth}
            \textbf{(f)}
        \end{minipage}%
        \hfill
        \begin{subfigure}[c]{0.100\linewidth}
            \centering
            \includegraphics[width=\linewidth]{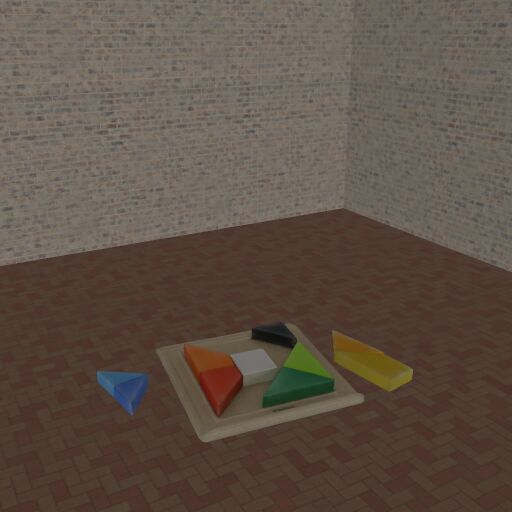}
        \end{subfigure}
        \hfill
        \begin{subfigure}[c]{0.100\linewidth}
            \centering
            \includegraphics[width=\linewidth]{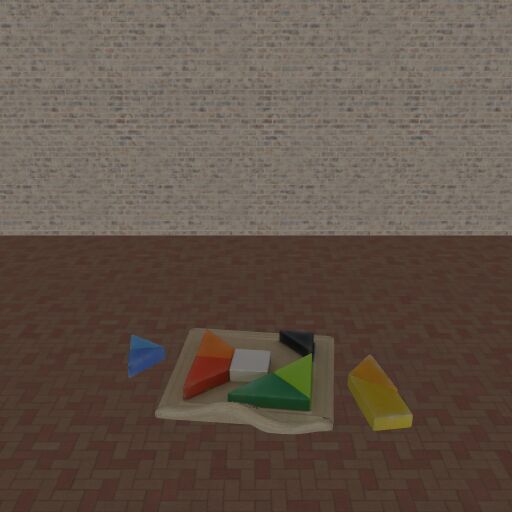}
        \end{subfigure}
        \hfill
        % ========================================================================================================================================
        \begin{subfigure}[c]{0.100\linewidth}
            \centering
            \includegraphics[width=\linewidth]{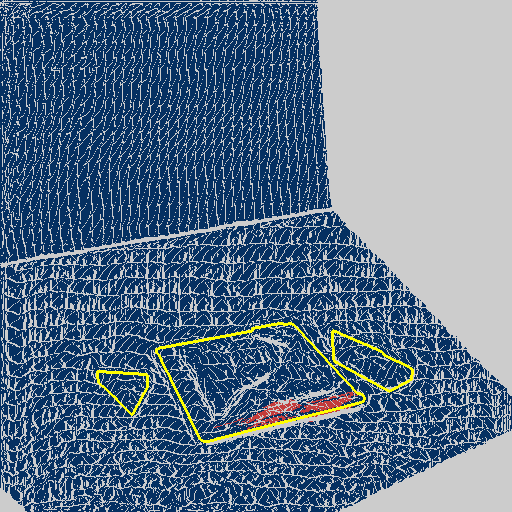}
        \end{subfigure}
        \hfill
        \begin{subfigure}[c]{0.100\linewidth}
            \centering
            \includegraphics[width=\linewidth]{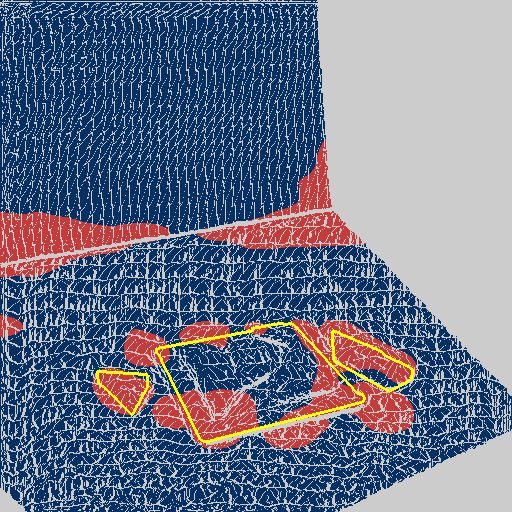}
        \end{subfigure}
        \hfill
        \begin{subfigure}[c]{0.100\linewidth}
            \centering
            \includegraphics[width=\linewidth]{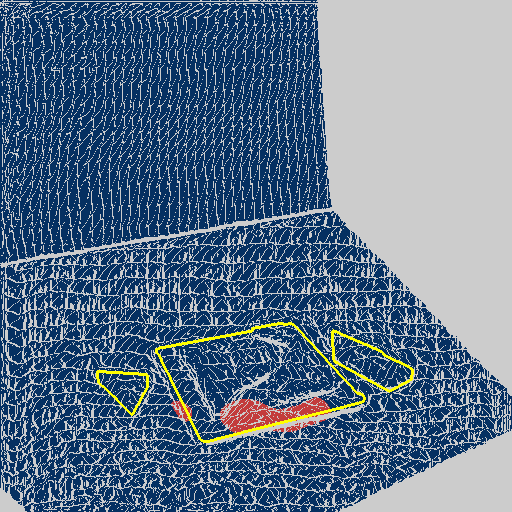}
        \end{subfigure}
        \hfill
        \begin{subfigure}[c]{0.100\linewidth}
            \centering
            \includegraphics[width=\linewidth]{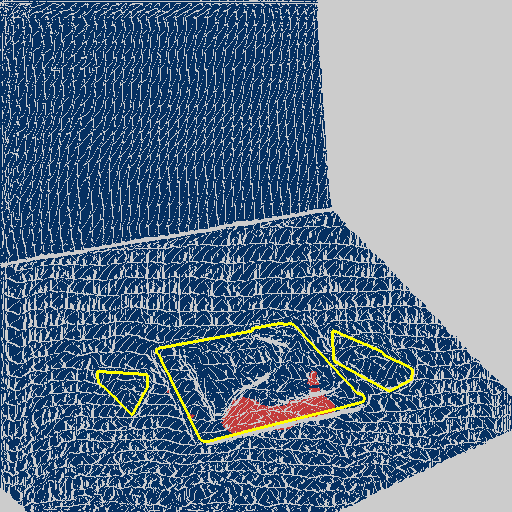}
        \end{subfigure}
        \hfill
        % ========================================================================================================================================
        \begin{subfigure}[c]{0.100\linewidth}
            \centering
            \includegraphics[width=\linewidth]{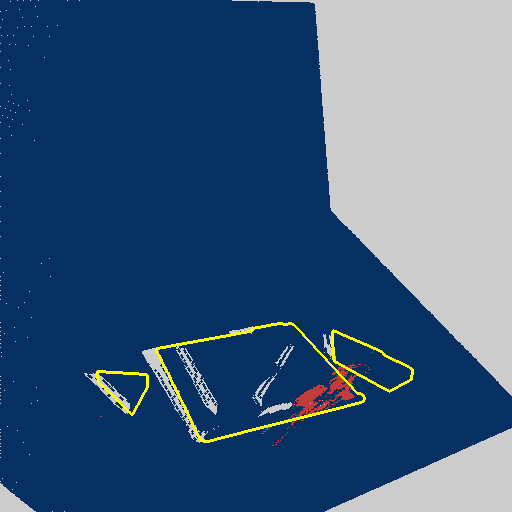}
        \end{subfigure}
        \hfill
        \begin{subfigure}[c]{0.100\linewidth}
            \centering
            \includegraphics[width=\linewidth]{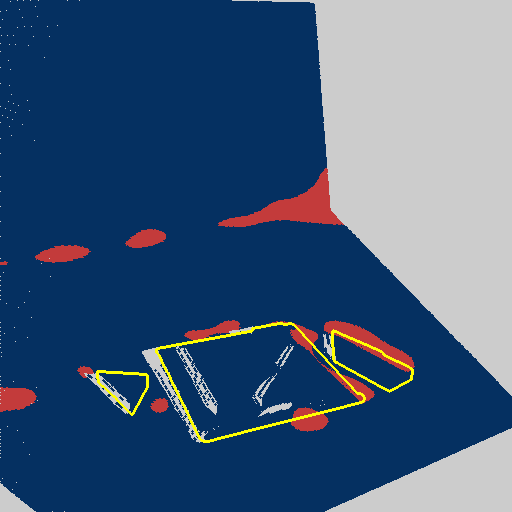}
        \end{subfigure}
        \hfill
        \begin{subfigure}[c]{0.100\linewidth}
            \centering
            \includegraphics[width=\linewidth]{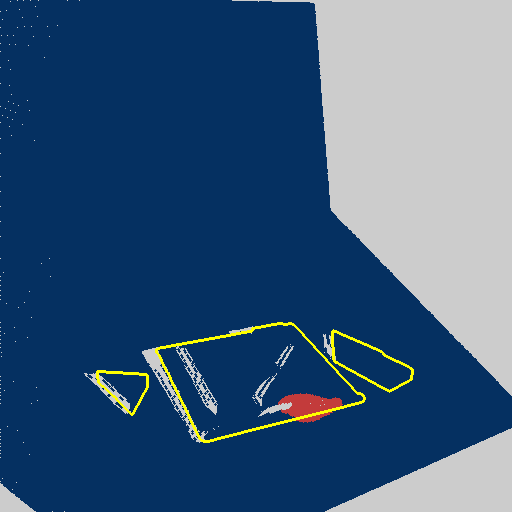}
        \end{subfigure}

        \vspace{0.3em} % Add vertical spacing between labels and images
    }

    % -----------------------------------------
    %        ROW 12 (DEF-DEF) (pair_03517)
    % -----------------------------------------
    
    {
        \begin{minipage}[c]{0.025\linewidth}
            \textbf{(g)}
        \end{minipage}%
        \hfill
        \begin{subfigure}[c]{0.100\linewidth}
            \centering
            \includegraphics[width=\linewidth]{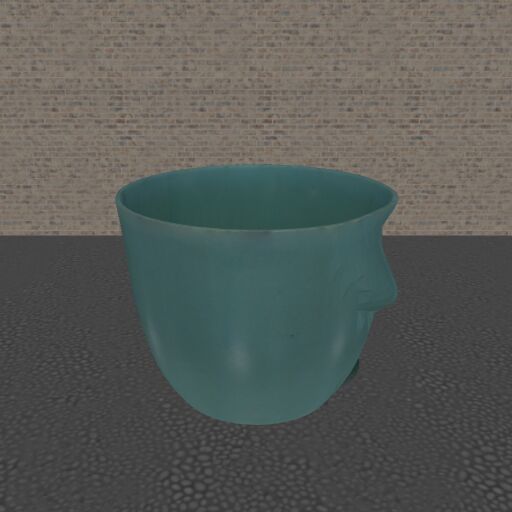}
        \end{subfigure}
        \hfill
        \begin{subfigure}[c]{0.100\linewidth}
            \centering
            \includegraphics[width=\linewidth]{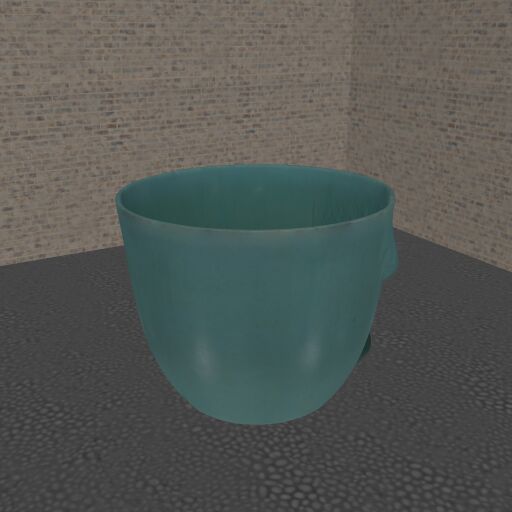}
        \end{subfigure}
        \hfill
        % ========================================================================================================================================
        \begin{subfigure}[c]{0.100\linewidth}
            \centering
            \includegraphics[width=\linewidth]{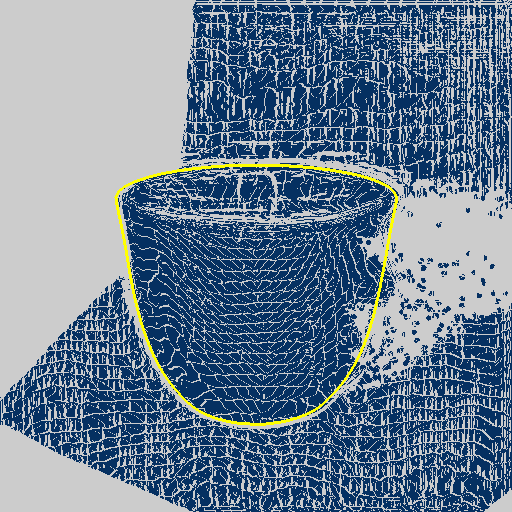}
        \end{subfigure}
        \hfill
        \begin{subfigure}[c]{0.100\linewidth}
            \centering
            \includegraphics[width=\linewidth]{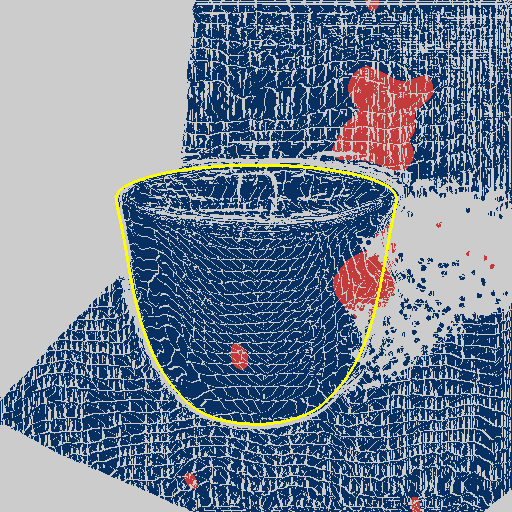}
        \end{subfigure}
        \hfill
        \begin{subfigure}[c]{0.100\linewidth}
            \centering
            \includegraphics[width=\linewidth]{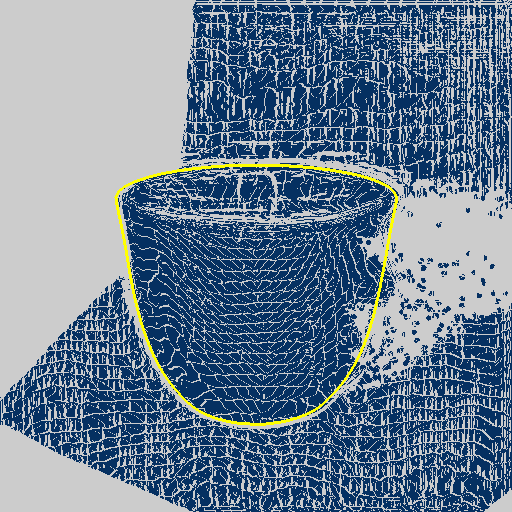}
        \end{subfigure}
        \hfill
        \begin{subfigure}[c]{0.100\linewidth}
            \centering
            \includegraphics[width=\linewidth]{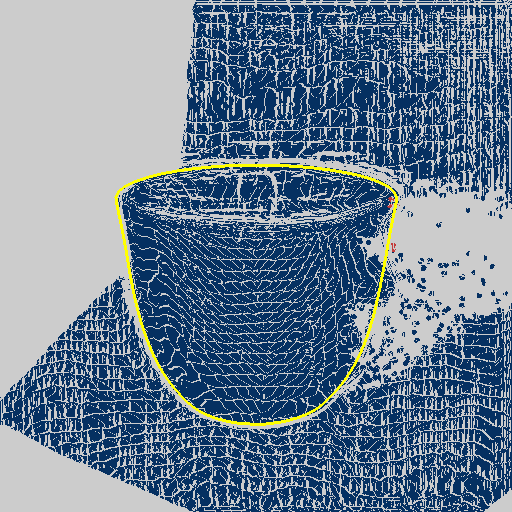}
        \end{subfigure}
        \hfill
        % ========================================================================================================================================
        \begin{subfigure}[c]{0.100\linewidth}
            \centering
            \includegraphics[width=\linewidth]{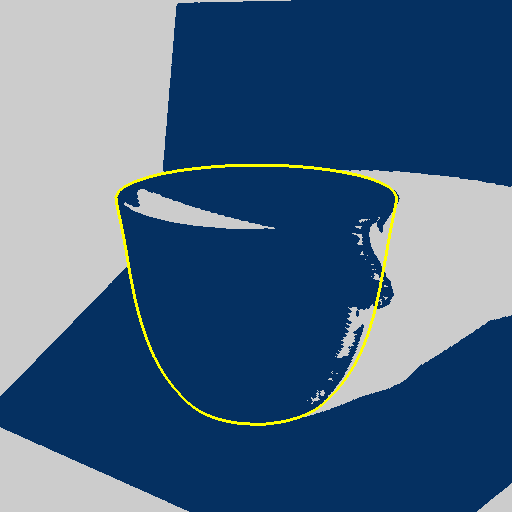}
        \end{subfigure}
        \hfill
        \begin{subfigure}[c]{0.100\linewidth}
            \centering
            \includegraphics[width=\linewidth]{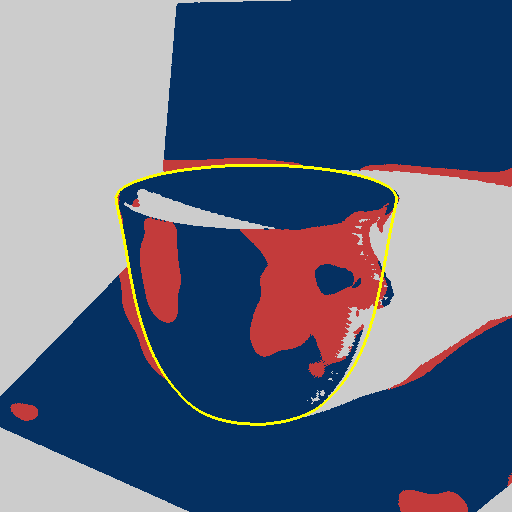}
        \end{subfigure}
        \hfill
        \begin{subfigure}[c]{0.100\linewidth}
            \centering
            \includegraphics[width=\linewidth]{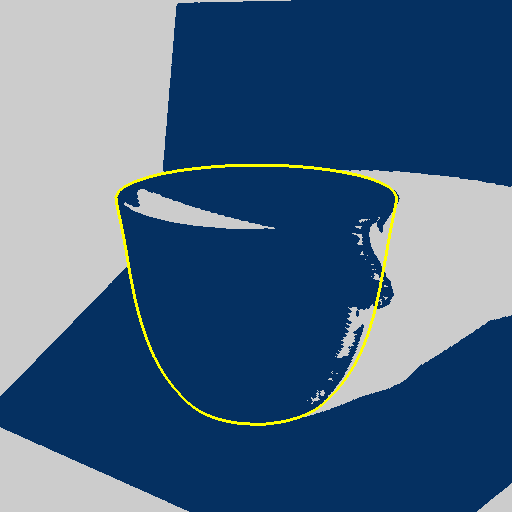}
        \end{subfigure}

        \vspace{0.3em} % Add vertical spacing between labels and images
    }

    % -----------------------------------------
    %        ROW 14 (DEF-DEF) (pair_04212)
    % -----------------------------------------

    {
        \begin{minipage}[c]{0.025\linewidth}
            \textbf{(h)}
        \end{minipage}%
        \hfill
        \begin{subfigure}[c]{0.100\linewidth}
            \centering
            \includegraphics[width=\linewidth]{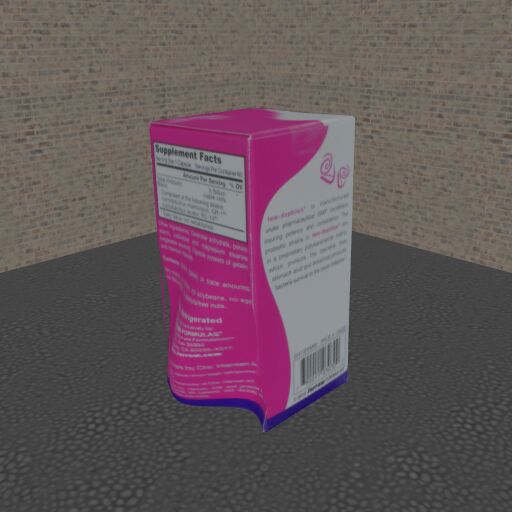}
        \end{subfigure}
        \hfill
        \begin{subfigure}[c]{0.100\linewidth}
            \centering
            \includegraphics[width=\linewidth]{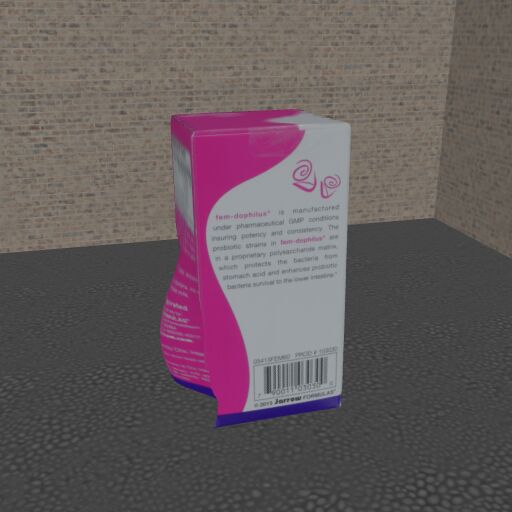}
        \end{subfigure}
        \hfill
        % ========================================================================================================================================
        \begin{subfigure}[c]{0.100\linewidth}
            \centering
            \includegraphics[width=\linewidth]{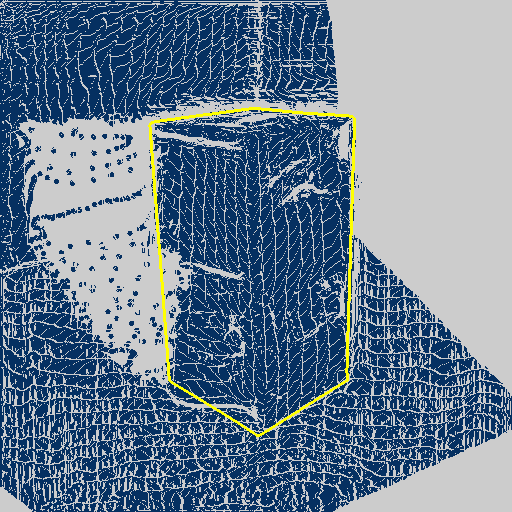}
        \end{subfigure}
        \hfill
        \begin{subfigure}[c]{0.100\linewidth}
            \centering
            \includegraphics[width=\linewidth]{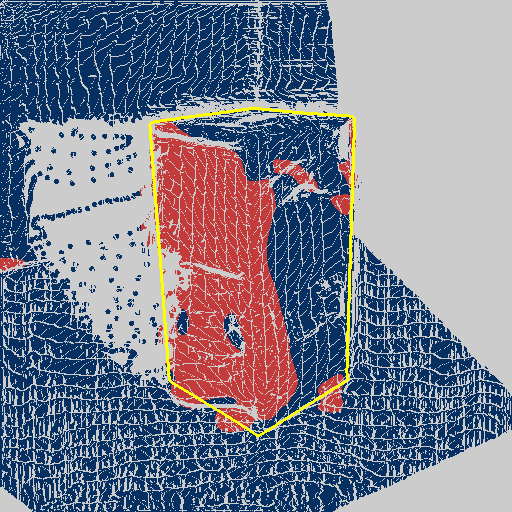}
        \end{subfigure}
        \hfill
        \begin{subfigure}[c]{0.100\linewidth}
            \centering
            \includegraphics[width=\linewidth]{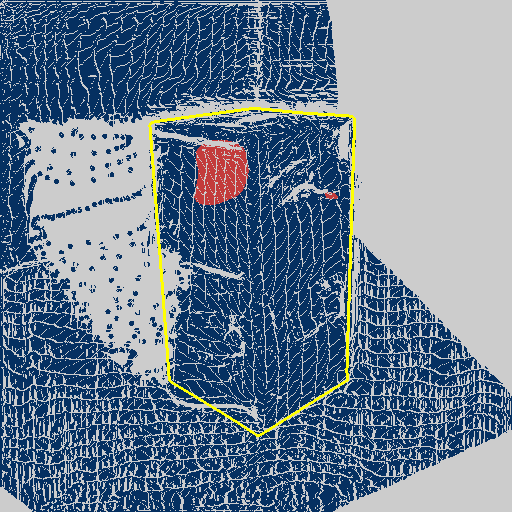}
        \end{subfigure}
        \hfill
        \begin{subfigure}[c]{0.100\linewidth}
            \centering
            \includegraphics[width=\linewidth]{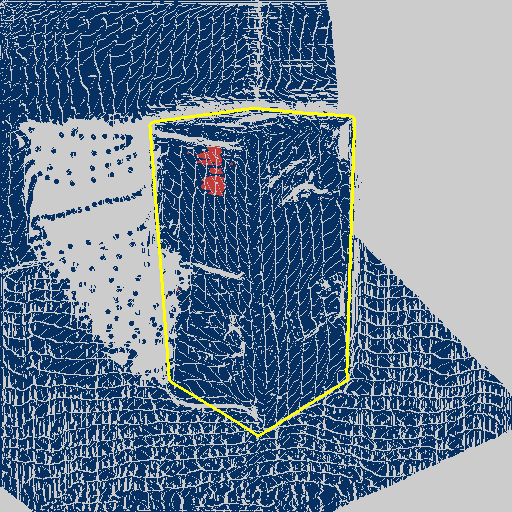}
        \end{subfigure}
        \hfill
        % ========================================================================================================================================
        \begin{subfigure}[c]{0.100\linewidth}
            \centering
            \includegraphics[width=\linewidth]{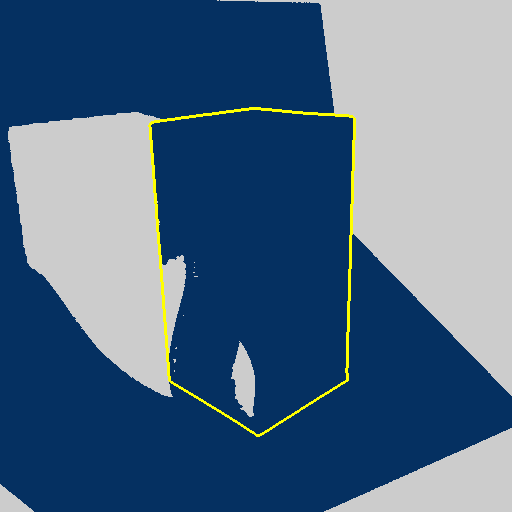}
        \end{subfigure}
        \hfill
        \begin{subfigure}[c]{0.100\linewidth}
            \centering
            \includegraphics[width=\linewidth]{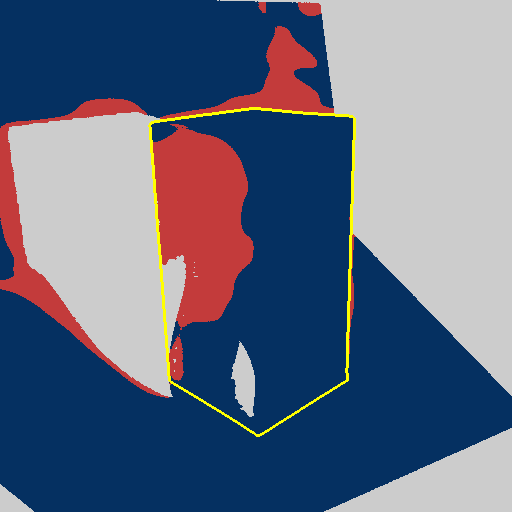}
        \end{subfigure}
        \hfill
        \begin{subfigure}[c]{0.100\linewidth}
            \centering
            \includegraphics[width=\linewidth]{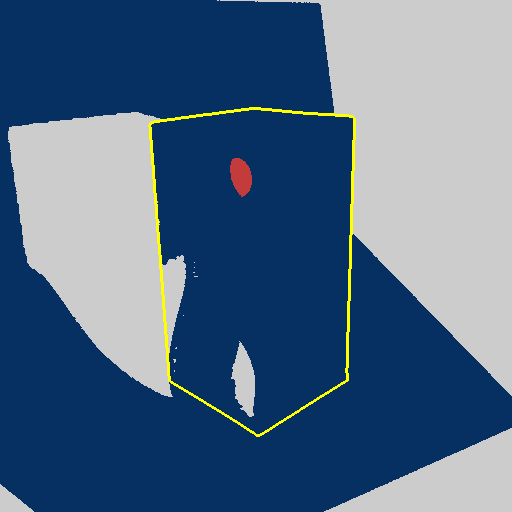}
        \end{subfigure}

        \vspace{0.3em} % Add vertical spacing between labels and images
    }

    % -----------------------------------------
    %        ROW 17 (DEF-DEF) (pair_05730)
    % -----------------------------------------

    {
        \begin{minipage}[c]{0.025\linewidth}
            \textbf{(i)}
        \end{minipage}%
        \hfill
        \begin{subfigure}[c]{0.100\linewidth}
            \centering
            \includegraphics[width=\linewidth]{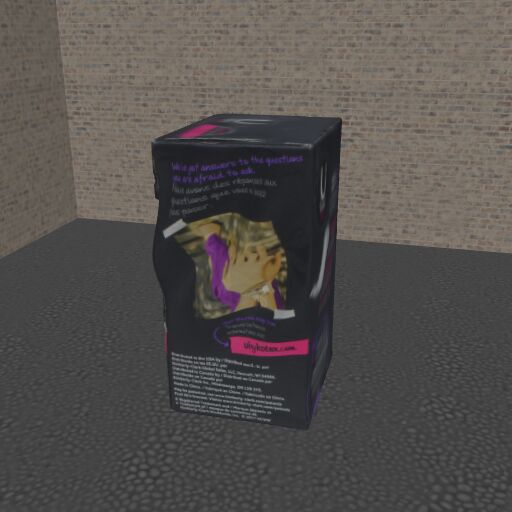}
        \end{subfigure}
        \hfill
        \begin{subfigure}[c]{0.100\linewidth}
            \centering
            \includegraphics[width=\linewidth]{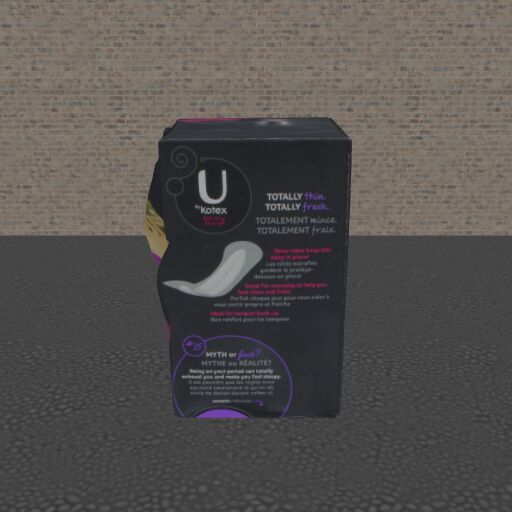}
        \end{subfigure}
        \hfill
        % ========================================================================================================================================
        \begin{subfigure}[c]{0.100\linewidth}
            \centering
            \includegraphics[width=\linewidth]{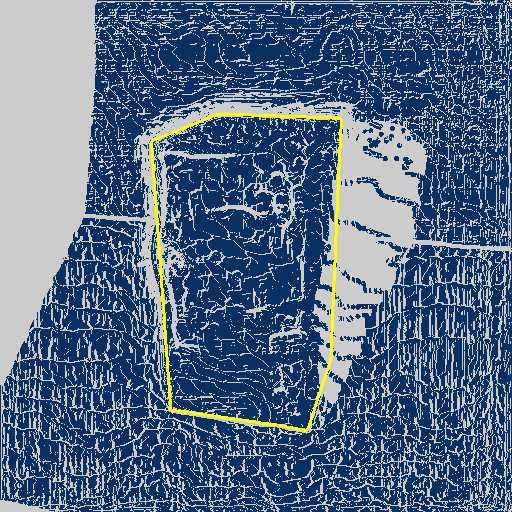}
        \end{subfigure}
        \hfill
        \begin{subfigure}[c]{0.100\linewidth}
            \centering
            \includegraphics[width=\linewidth]{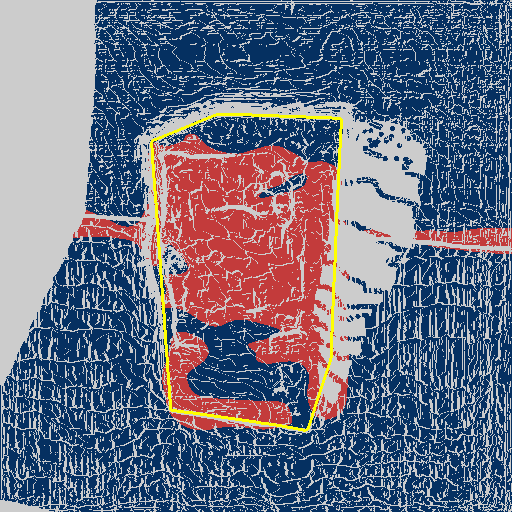}
        \end{subfigure}
        \hfill
        \begin{subfigure}[c]{0.100\linewidth}
            \centering
            \includegraphics[width=\linewidth]{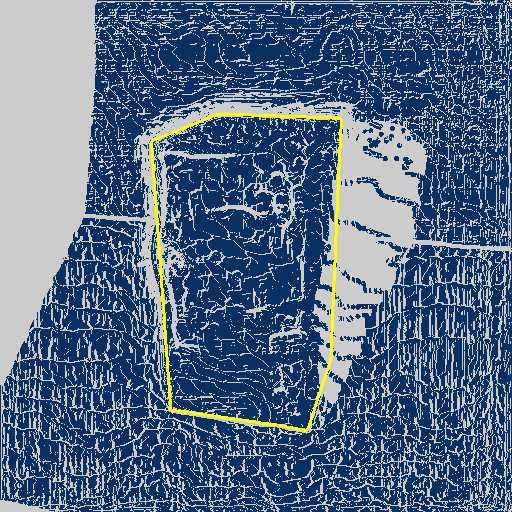}
        \end{subfigure}
        \hfill
        \begin{subfigure}[c]{0.100\linewidth}
            \centering
            \includegraphics[width=\linewidth]{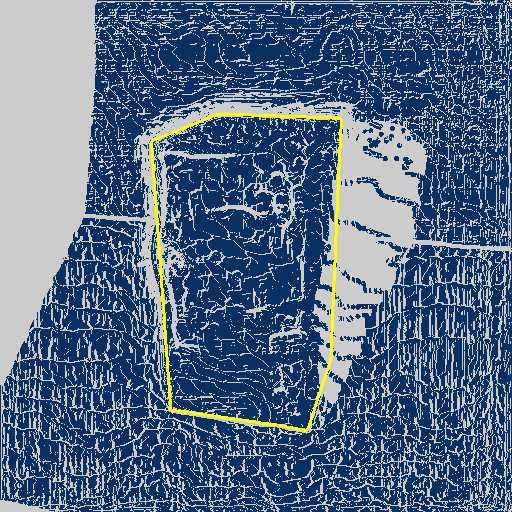}
        \end{subfigure}
        \hfill
        % ========================================================================================================================================
        \begin{subfigure}[c]{0.100\linewidth}
            \centering
            \includegraphics[width=\linewidth]{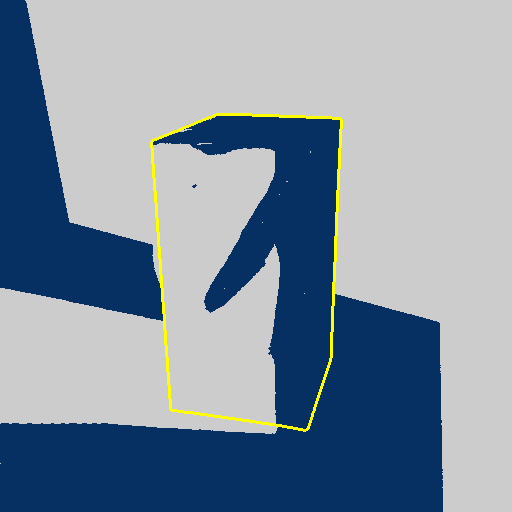}
        \end{subfigure}
        \hfill
        \begin{subfigure}[c]{0.100\linewidth}
            \centering
            \includegraphics[width=\linewidth]{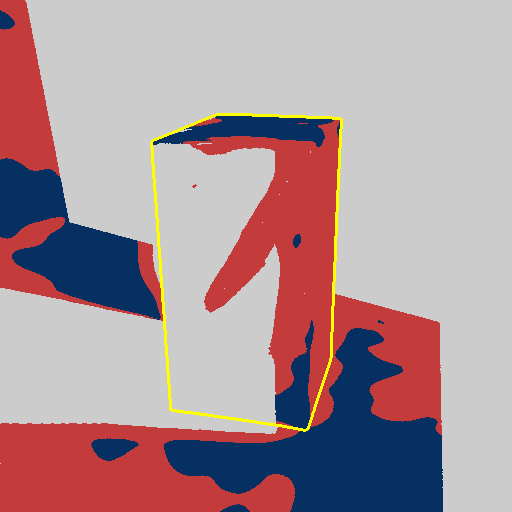}
        \end{subfigure}
        \hfill
        \begin{subfigure}[c]{0.100\linewidth}
            \centering
            \includegraphics[width=\linewidth]{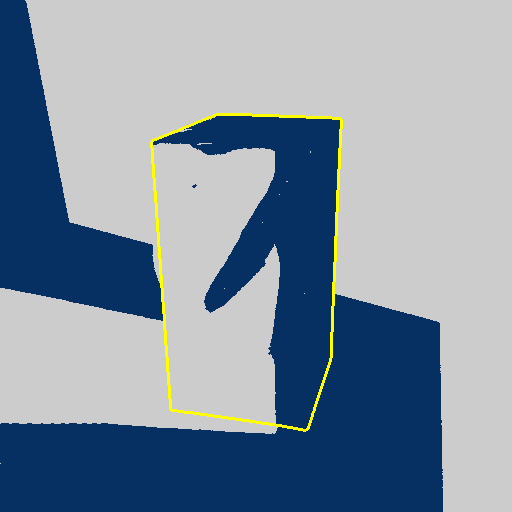}
        \end{subfigure}

        % \vspace{-0.6em}
    }
    
    \caption{
        Image pairs from the DeformView test set and their corresponding predictions by all evaluated methods. 
        Rows (a)--(c) show consistent \textit{original-original} pairs, rows (d)--(f) inconsistent \textit{original-deformed} pairs, and rows (g)--(i) consistent \textit{deformed-deformed} pairs.
        Predictions are shown from the perspective of view~1, with the object outline indicated in yellow.
        The examples highlight common failure modes, including reprojection errors and visually unusual yet geometrically consistent regions, and show how DEFECt3R suppresses many resulting false positives.
    }

    \Description{
        A large grid of qualitative examples from the DeformView test set. The figure is organized into nine rows grouped into three categories: consistent original-original pairs, inconsistent original-deformed pairs, and consistent deformed-deformed pairs. Each row begins with two input views of the same object. The remaining columns show ground-truth inconsistency masks and predictions produced by MEt3R and DEFECt3R, using both MASt3R-estimated poses and ground-truth poses. Predictions are visualized as heat maps, where brighter regions indicate higher inconsistency scores, and object boundaries are marked in yellow. The examples illustrate false positives caused by pose-estimation errors and visually unusual but geometrically consistent regions, as well as the improved localization produced by DEFECt3R.
    }
    
    \label{figure:evaluation-full}
    
\end{figure*}

%% file: acknowledgments.tex
\begin{acks}
This work was funded in part by IDLab (Ghent University -- imec), by Flanders Innovation \& Entrepreneurship (VLAIO), by Research Foundation -- Flanders (FWO) (1SA9O26N \& G0A2523N), and by the European Union. 
\end{acks}

%% file: references.bib
@InProceedings{Imagen,
    title       = {Photorealistic Text-to-Image Diffusion Models with Deep Language Understanding},
    author      = {Saharia, Chitwan and Chan, William and Saxena, Saurabh and Li, Lala and Whang, Jay and Denton, Emily L and Ghasemipour, Kamyar and Gontijo Lopes, Raphael and Karagol Ayan, Burcu and Salimans, Tim and Ho, Jonathan and Fleet, David and Norouzi, Mohammad},
    booktitle = {NeurIPS},
    pages       = {36479--36494},
    publisher   = {Curran Associates, Inc.},
    volume      = {35},
    year        = {2022}
}

@article{Glide,
    title={{Glide}: Towards photorealistic image generation and editing with text-guided diffusion models},
    author={Nichol, Alex and Dhariwal, Prafulla and Ramesh, Aditya and Shyam, Pranav and Mishkin, Pamela and McGrew, Bob and Sutskever, Ilya and Chen, Mark},
    journal={arXiv preprint arXiv:2112.10741},
    year={2021}
}

@inproceedings{GenWarp,
    title       = {{GenWarp}: Single Image to Novel Views with Semantic-Preserving Generative Warping},
    author      = {Seo, Junyoung and Fukuda, Kazumi and Shibuya, Takashi and Narihira, Takuya and Murata, Naoki and Hu, Shoukang and Lai, Chieh-Hsin and Kim, Seungryong and Mitsufuji, Yuki},
    booktitle   = {NeurIPS},
    pages       = {80220--80243},
    publisher   = {Curran Associates, Inc.},
    volume      = {37},
    year        = {2024},
    doi         = {10.52202/079017-2550},
}

@inproceedings{elata2025novel,
  title={Novel view synthesis with pixel-space diffusion models},
  author={Elata, Noam and Kawar, Bahjat and Ostrovsky-Berman, Yaron and Farber, Miriam and Sokolovsky, Ron},
  booktitle={CVPR},
  pages={26756--26766},
  year={2025}
}

@INPROCEEDINGS{rombach2021geometry,
  title={Geometry-Free View Synthesis: Transformers and no 3D Priors}, 
  author={Rombach, Robin and Esser, Patrick and Ommer, Björn},
  booktitle={ICCV}, 
  pages={14336--14346},
  year={2021},
  doi={10.1109/ICCV48922.2021.01409}
}

@ARTICLE{overview-SID,
    title       ={Synthetic Image Verification in the Era of Generative Artificial Intelligence: What Works and What Isn't There yet}, 
    author      ={Tariang, Diangarti and Corvi, Riccardo and Cozzolino, Davide and Poggi, Giovanni and Nagano, Koki and Verdoliva, Luisa},
    journal     ={IEEE Security \& Privacy}, 
    year        ={2024},
    volume      ={22},
    number      ={3},
    pages       ={37--49},
    doi         ={10.1109/MSEC.2024.3376637}
}

@InProceedings{NPR,
    title       = {Rethinking the Up-Sampling Operations in CNN-based Generative Network for Generalizable Deepfake Detection},
    author      = {Tan, Chuangchuang and Zhao, Yao and Wei, Shikui and Gu, Guanghua and Liu, Ping and Wei, Yunchao},
    booktitle   = {CVPR},
    month       = {June},
    year        = {2024},
    pages       = {28130--28139}
}

@InProceedings{CNN-DCT,
  title         = {Leveraging Frequency Analysis for Deep Fake Image Recognition},
  author        = {Frank, Joel and Eisenhofer, Thorsten and Sch{\"o}nherr, Lea and Fischer, Asja and Kolossa, Dorothea and Holz, Thorsten},
  booktitle     = {ICML},
  pages         = {3247--3258},
  year          = {2020},
  volume        = {119},
  series        = {Proceedings of Machine Learning Research},
  month         = {13--18 Jul},
  publisher     = {PMLR},
}

@InProceedings{DnCNN-SID,
    author      = {Corvi, Riccardo and Cozzolino, Davide and Poggi, Giovanni and Nagano, Koki and Verdoliva, Luisa},
    title       = {Intriguing Properties of Synthetic Images: From Generative Adversarial Networks to Diffusion Models},
    booktitle   = {CVPR},
    month       = {June},
    year        = {2023},
    pages       = {973--982}
}

@InProceedings{UnivFD,
    author      = {Ojha, Utkarsh and Li, Yuheng and Lee, Yong Jae},
    title       = {Towards Universal Fake Image Detectors That Generalize Across Generative Models},
    booktitle   = {CVPR},
    month       = {June},
    year        = {2023},
    pages       = {24480--24489}
}

@InProceedings{ShadowLines,
    author          = {Sarkar, Ayush and Mai, Hanlin and Mahapatra, Amitabh and Lazebnik, Svetlana and Forsyth, D.A. and Bhattad, Anand},
    title           = {Shadows Don't Lie and Lines Can't Bend! Generative Models don't know Projective Geometry...for now},
    booktitle       = {CVPR},
    month           = {June},
    year            = {2024},
    pages           = {28140--28149}
}

@article{3D-consistency-scoring,
    title           ={Novel view synthesis with diffusion models},
    author          ={Watson, Daniel and Chan, William and Martin-Brualla, Ricardo and Ho, Jonathan and Tagliasacchi, Andrea and Norouzi, Mohammad},
    journal         ={arXiv preprint arXiv:2210.04628},
    year            ={2022}
}

@InProceedings{TSED,
    author          = {Yu, Jason J. and Forghani, Fereshteh and Derpanis, Konstantinos G. and Brubaker, Marcus A.},
    title           = {Long-Term Photometric Consistent Novel View Synthesis with Diffusion Models},
    booktitle       = {ICCV},
    month           = {October},
    year            = {2023},
    pages           = {7094--7104}
}

@InProceedings{MEt3R,
    title           = {{MET3R}: Measuring Multi-View Consistency in Generated Images},
    author          = {Asim, Mohammad and Wewer, Christopher and Wimmer, Thomas and Schiele, Bernt and Lenssen, Jan Eric},
    booktitle       = {CVPR},
    month           = {June},
    year            = {2025},
    pages           = {6034--6044}
}

@misc{GeCo,
      title={GeCo: Evaluating Geometric Consistency for Video Generation via Motion and Structure}, 
      author={Leslie Gu and Junhwa Hur and Charles Herrmann and Fangneng Zhan and Todd Zickler and Deqing Sun and Hanspeter Pfister},
      year={2026},
      eprint={2512.22274},
      archivePrefix={arXiv},
      primaryClass={cs.CV},
      url={https://arxiv.org/abs/2512.22274}, 
}

@article{NeRF,
    title           = {{NeRF}: representing scenes as neural radiance fields for view synthesis},
    author          = {Mildenhall, Ben and Srinivasan, Pratul P. and Tancik, Matthew and Barron, Jonathan T. and Ramamoorthi, Ravi and Ng, Ren},
    year            = {2021},
    issue_date      = {January 2022},
    publisher       = {Association for Computing Machinery},
    address         = {New York, NY, USA},
    volume          = {65},
    number          = {1},
    issn            = {0001-0782},
    doi             = {10.1145/3503250},
    journal         = {Commun. ACM},
    month           = dec,
    pages           = {99--106},
    numpages        = {8}
}

@InProceedings{MASt3R,
    title           = {Grounding Image Matching in 3D with MASt3R},
    author          = {Leroy, Vincent and Cabon, Yohann and Revaud, Jerome},
    booktitle       = {ECCV},
    year            = {2025},
    publisher       = {Springer Nature Switzerland},
    address         = {Cham},
    pages           = {71--91},
    isbn            = {978-3-031-73220-1}
}

@InProceedings{Google_SO,
    title           = {Google Scanned Objects: A High-Quality Dataset of 3D Scanned Household Items}, 
    author          = {Downs, Laura and Francis, Anthony and Koenig, Nate and Kinman, Brandon and Hickman, Ryan and Reymann, Krista and McHugh, Thomas B. and Vanhoucke, Vincent},
    booktitle       = {ICRA}, 
    year            = {2022},
    pages           = {2553--2560},
    doi             = {10.1109/ICRA46639.2022.9811809}
}

@article{object-level-inconsistency-dataset,
    title           = {Multimodal Language Models Cannot Spot Spatial Inconsistencies},
    author          = {Khangaonkar, Om and Rad, Hadi J and Pirsiavash, Hamed},
    journal         = {arXiv preprint arXiv:2604.00799},
    year            = {2026}
}

@InProceedings{DINO,
    title           = {Emerging Properties in Self-Supervised Vision Transformers},
    author          = {Caron, Mathilde and Touvron, Hugo and Misra, Ishan and J\'egou, Herv\'e and Mairal, Julien and Bojanowski, Piotr and Joulin, Armand},
    booktitle       = {ICCV},
    month           = {October},
    year            = {2021},
    pages           = {9650--9660}
}

@inproceedings{FeatUp,
    title           = {{FeatUp}: A Model-Agnostic Framework for Features at Any Resolution},
    author          = {Fu, Stephanie and Hamilton, Mark and Brandt, Laura E. and Feldmann, Axel and Zhang, Zhoutong and Freeman, William},
    booktitle       = {ICLR},
    pages           = {45324--45350},
    volume          = {2024},
    year            = {2024}
}

@INPROCEEDINGS{Zhang2009TwoView,
  author        ={Zhang, Wei and Cao, Xiaochun and Feng, Zhiyong and Zhang, Jiawan and Wang, Ping},
  booktitle     ={ICME}, 
  title         ={Detecting photographic composites using two-view geometrical constraints}, 
  year          ={2009},
  volume        ={},
  number        ={},
  pages         ={1078--1081},
  doi={10.1109/ICME.2009.5202685}
}
